%% file: main.tex
\documentclass[sigconf]{acmart}
\usepackage{float}
\usepackage{booktabs}
\usepackage{textcomp}
\usepackage{graphicx}
\usepackage{ragged2e}
\usepackage{booktabs} % For formal tables
\usepackage[colorinlistoftodos]{todonotes}

\providecommand{\tabularnewline}{\\}
\floatstyle{ruled}
\newfloat{algorithm}{tbp}{loa}
\providecommand{\algorithmname}{Algorithm}
\floatname{algorithm}{\protect\algorithmname}

\newcounter{todocounter}

\begin{document}
\title{Batch Tournament Selection for Genetic Programming}

\subtitle{The quality of lexicase, the speed of Tournament}

%\titlenote{Produces the permission block, and
%  copyright information}
%\titlenote{}

%\subtitlenote{The full version of the author's guide is available as
%  \texttt{acmart.pdf} document}

\author{Vin\'{i}cius V. de Melo}
\affiliation{%
	\institution{Data Science Team\\
		SkipTheDishes Restaurant Services Inc.}
	\city{Winnipeg}
	\state{MB, Canada}
}
\email{vinicius.melo@skipthedishes.ca}

\author{Danilo Vasconcellos Vargas}
\affiliation{%
	\institution{Faculty of Information Science and Electrical Engineering \\ Kyushu University}
	\streetaddress{Address}
	\city{Fukuoka}
	\state{Japan}
	\postcode{postcode}
}
\email{vargas@inf.kyushu-u.ac.jp}

\author{Wolfgang Banzhaf}
\affiliation{%
	\institution{CSE Dept \& BEACON Center \\ Michigan State University}
	\streetaddress{Address}
	\city{East Lansing, MI}
	\state{USA}
	\postcode{48864}
}
\email{banzhafw@msu.edu}

% The default list of authors is too long for headers.
\renewcommand{\shortauthors}{V. V. de Melo et al.}

\begin{abstract}
Lexicase selection achieves very good solution quality  by introducing ordered test cases.
However, the computational complexity of lexicase selection can prohibit its use in many applications.
In this paper, we introduce Batch Tournament Selection (BTS), a hybrid of tournament and lexicase selection which is approximately one order of magnitude faster than lexicase selection while achieving a competitive quality of solutions.
Tests on a number of regression datasets show that BTS compares well with lexicase selection in terms of mean absolute error while having a speed-up of up to $25$ times.
Surprisingly, BTS and lexicase selection have almost no difference in both diversity and performance.
This reveals that batches and ordered test cases are completely different mechanisms which share the same general principle fostering the specialization of individuals.
This work introduces an efficient algorithm that sheds light onto the main principles behind the success of lexicase, potentially opening up a new range of possibilities for algorithms to come.
\end{abstract}

%
% The code below should be generated by the tool at
% http://dl.acm.org/ccs.cfm
% Please copy and paste the code instead of the example below. 
%



\copyrightyear{2019} 
\acmYear{2019} 
\setcopyright{acmcopyright}
\acmConference[GECCO '19]{Genetic and Evolutionary Computation Conference}{July 13--17, 2019}{Prague, Czech Republic}
\acmBooktitle{Genetic and Evolutionary Computation Conference (GECCO '19), July 13--17, 2019, Prague, Czech Republic}
\acmPrice{15.00}
\acmDOI{10.1145/3321707.3321793}
\acmISBN{978-1-4503-6111-8/19/07}

\maketitle

\input{body_contents}

% \clearpage
\bibliographystyle{ACM-Reference-Format}

% \nocite{*}
\bibliography{references/referencias,selection}

\end{document}

%% file: body_contents.tex
% \vinicius{Must change the CCS Concepts}

\section{Introduction}

In traditional Genetic Programming and similar algorithms, the fitness
of an individual is quantified as its aggregate performance over the
test cases of a training set. The aggregation produces a quality measure
such as \textit{Mean Absolute Error} (MAE),   \textit{Correlation Coefficient}, or a \textit{Loss Function}. In
order to choose parents for reproduction or to truncate the population for survival into the next generation selection methods such as
roulette wheel or tournament selection employ this aggregated fitness measure for deciding
which individuals are worth reproducing. 
However, lumping together the overall behavior of a program or algorithm into one number -- despite it arising from probably thousands of fitness cases -- has been found to throw away valuable information \cite{krawiec2015automatic}. 

Over the years, a number of approaches have been tried to make better use of individual fitness cases. The most extreme suggestion was to use a randomly chosen fitness case on a randomly chosen individual for evaluation \cite{nordin1997line}, assuming that a useful fitness measure emerges by averaging over the long run. This extreme approach was used in cases where fitness evaluation takes a long time, like in robotics, rendering evaluation otherwise close to impossible.   
Other effective approaches were suggested, too, like creating sub-populations for each similar input which results in the concept of niched fitness (multiple fitness functions divided in sub-populations) \cite{vargas2013aself,Oltean03acomparison},   replacing the test cases by a model function~\cite{nordin1998evolution}, and more generally in EC, to use a surrogate fitness function~\cite{jin2011surrogate}.

A few years ago, \textit{lexicase selection}  \cite{spector2012assessment} was proposed as an alternative that also considers all fitness cases. Lexicase selection works by testing on fitness cases
separately but systematically through simply changing the order of their application, thus giving individuals that  behave well on particular fitness cases, but not in general, a chance to have offspring. Lexicase selection
works nicely for discrete problems, while $\epsilon$-Lexicase \cite{la2016epsilon}
has been recently proposed for continuous problems.

Although lexicase parent selection improves
the search, it demands very high computational efforts because
all individuals in the population must be compared on a large number
of fitness cases. These numbers shrink as it advances in the fitness cases, but it
may take a while to select a single parent. Following lexicase's approach,
we propose the Batch Tournament Selection. It splits  fitness cases
into batches and runs regular tournament selection on them. A batch
of size one is equivalent to lexicase selection, while a batch containing all cases is equivalent to a canonical tournament selection algorithm. By tuning the batch size parameter, one may approximate
the quality of Lexicase with the speed of  canonical tournaments.

One of the important aspects of lexicase selection is its success in keeping a population diverse over repeated rounds of selection. As well, since the selection pressure is constantly changing, lexicase selection helps to avoid overfitting and produces better solutions faster. There are also problems with Lexicase as when it does ''hyperselection'', i.e. when the population is reduced to just one niche, thus getting stuck on the way to an optimal solution.

Here we want to address the issues of speed and diversity and propose an algorithm that combines strong features of tournament selection with those of Lexicase in our algorithm called batch tournament selection (BTS). 

The paper is organized as follows: Section 2 discusses selection algorithms, Section 3 introduces batch tournament selection and its variant, Batch Tournament Selection Shuffled (BTSS). Section 4 reports our results and Section 5 presents our conclusions.

\section{\label{sec:Selection-algos}Selection algorithms}

Evolution needs selection to bring forth better candidate solutions.
Throughout evolution, these better candidate solutions will help generate even better ones.
This will drive evolution towards candidate solutions with high fitness.
However, the process of generating better candidate solutions is constrained by time as well as the need for an important ingredient, diversity.
Time-consuming selection procedures will often cause a slow-down, preventing runs from achieving satisfactory results in a timely manner and prohibiting their use in real time applications.
Moreover, populations that lack diversity will consequently fail to explore the fitness landscape.

Selection pressure and diversity are opposing forces in evolution with time being the overall constraint that often allows simple/approximate solutions to thrive over complex/exact ones.
Depending on the selection method chosen, a completely different trade-off between selection pressure, diversity and time takes place.
%However, even when diversity is equally preserved, not all selections are the same.
Proportionate reproduction (roulette) was one of the first proposed selection methods.
It is not easy to control selection pressure in roulette wheel selection, but the time complexity when efficiently coded can achieve $O(N*T)$ for population of size $N$ and $T$ test cases.

Tournament selection is another widely known method which selects individuals based on tournament winners.
Since weaker individuals have less chance to win in bigger tournaments, the tournament size provides the selection pressure \cite{miller1995genetic}.
Tournament selection is a relatively fast method with time complexity $O(N*T)$.
It can be easily parallelized and has a selection pressure that can be easily adjusted \cite{goldberg1991comparative}.

Recently, lexicase selection was proposed.
It selects individuals that perform well in random test sequences. 
In this manner, the variance of the sequence distribution provides a good trade-off between diversity and selection pressure \cite{spector2012assessment}.
Interestingly, random test sequences force candidate solutions to solve small portions of the problem well and therefore tends to select candidate solutions which are specialists.
This is in stark contrast with the widespread use of average of solutions which selects for generalists.
In fact, Lexicase has been shown to outperform other methods in many different scenarios \cite{helmuth2015solving, helmuth2015general, liskowski2015comparison}.
A big disadvantage, however, lies in its time complexity which is $O(N^2*T)$.

% Random, Roulette, Tournament, etc

%\vinicius{ 
La Cava et. al~\cite{la2016epsilon} introduced $\epsilon$-Lexicase, a new form of lexicase selection for symbolic regression in the continuous domain, whereas Lexicase was originally proposed for discrete problems. It was compared
with tournament selection and age-fitness Pareto optimization on a series of regression data sets. Different versions of $\epsilon$-Lexicase were capable of producing fitter models
than the other methods. The authors also proposed a strategy to automatically adapt $\epsilon$ based on the population performance distribution, which produced a very small computational overhead while achieving high-quality results. This is the version employed as baseline in this paper, here named Ae-Lex.
%}

\section{Batch Tournament Selection}

In batch tournament selection (BTS) 
fitness cases are, in each generation, split into batches and all batches are processed.
BTS first orders the fitness cases by their error, so cases with a
similar difficulty level belong to the same batch. 
On the other hand, a second version of the algorithm, BTSS, shuffles the cases in addition to grouping them in batches.

For each batch, the method first calculates the fitness, such as MAE, of each individual in the population. 
It then follows with a tournament selection in which randomly selected 
individuals participate. In each tournament, the participating individual that
performs best on that particular batch survives to the next generation. 
The method then goes to the next batch, repeating the procedure, until all batches are processed. 
If the number of selected individuals is smaller than the population size then the process repeats with
new batches.

BTS and BTSS return a list of $k$ parents selected for mating. 
Considering a parent is selected from a single batch, only $k$ batches are needed. 
Thus, although all individuals must be evaluated in all their fitness cases, not all fitness cases may be considered in the comparison for selecting the $k$ parents. 
If the number of batches is smaller than $k$, a new round starts from the first batch to continue selecting  parents. 
Ae-Lex, on the other hand, may need to examine the entire data set if candidate parents are clones of each other and the duplicates are not removed. However, since Ae-Lex maintains a large population diversity, this cloning effect was observed only in a small part of the population.

Here, we tested two approaches. In the first one, BTS, we study a type of phenotypic clustering in which fitness cases are ordered by their difficulty level. 
To do so, the error of each fitness case is calculated (such as MAE, Mean Squared Error, or Root Mean Squared Error) and cases with similar difficulty are grouped in the same batch. 
To force BTS to select  individuals that perform best in the most difficult cases and let the selection of individuals for easy cases depend on the batch size, we use a descending order of difficulty for batches.
In other words, the first batch contains the hardest examples while the last batch contains the easiest ones. 
For a detailed description of BTS, please refer to Algorithm~\ref{bts_alg}.

The second approach, called BTSS, makes use of mixed difficulty test cases within batches.  
Fitness cases are permuted, not sampled with repetition. 
In this approach, individuals must perform well on all difficulty levels to be selected.
Thus, it results in a different type of selection pressure when compared with the heterogenous difficulty of batches from BTS. 
Since the examples within a batch are randomly selected, it might, however, happen that a batch has only easy cases allowing poor-quality individuals to be selected but that should be a rare occurrence.

\begin{algorithm*}
	\caption{\label{alg:A-simple-pseudo-code}A simple pseudocode for BTS. Since
		we are minimizing the error, we use \emph{min} to select the best
		individual. Collection indices start in 1.}
%	
%	\begin{tabular}{rp{5cm}}
%		\textbf{population} & the entire population\tabularnewline
%		\textbf{k} & the number of parents to be selected\tabularnewline
%		\textbf{batch\_size} & the size of each batch, i.e., the number of cases in each batch\tabularnewline
%		\textbf{tourn\_size} & the number of individuals to be selected for tournament\tabularnewline
%		\textbf{shuffle} & a boolean indicating whether the cases should be shuffled before creating
%		the batches\tabularnewline
%		%\tabularnewline
%		%\hline 
%		\textbf{selected\_individuals} & the collection of parents\tabularnewline
%		\hline 
%	\end{tabular}

\begin{tabular}{rrp{12.7cm}}
	& \textbf{population} & the entire population\tabularnewline
	& \textbf{k} & the number of parents to be selected\tabularnewline
	\textbf{Input} & \textbf{batch\_size} & the size of each batch, i.e., the number of cases in each batch\tabularnewline
	& \textbf{tourn\_size} & the number of individuals to be selected for tournament\tabularnewline
	& \textbf{shuffle} & a boolean indicating whether the cases should be shuffled before creating
	the batches\tabularnewline
	\hline 
	\textbf{Output } & \textbf{selected\_individuals} & the collection of parents\tabularnewline
	\hline 
\end{tabular}

    \small
     \begin{flushleft}
     	
    % ~\\
	~\\
	\textbf{function} selBatchTournament(\emph{population}, \emph{k},
	\emph{batch\_size}, \emph{tourn\_size}, \emph{shuffle})
	
% 	~

	\hspace{0.5cm}\textbf{if} shuffle \textbf{then}

	\hspace{0.5cm}\hspace{0.5cm}idx\_cases\_batch = 1,...,|population{[}1{]}.case\_error|  // \emph{An array of indexes: case\_error contains the error for each 	case}\\
	
	\hspace{0.5cm}\hspace{0.5cm}shuffle(idx\_cases\_batch)
	
	\hspace{0.5cm}\textbf{else}
	
	\hspace{0.5cm}\hspace{0.5cm}// \emph{Order by difficulty according to the fitness
		of the best solution so far} \\
	
	\hspace{0.5cm}\hspace{0.5cm}best = find\_best(population)
	
	\hspace{0.5cm}\hspace{0.5cm}idx\_cases\_batch = order(best.case\_error)   // \emph{The index of the cases in decreasing order (harder cases first)}
	
	\hspace{0.5cm}\textbf{end if}
	
	~
	
	\hspace{0.5cm}// \emph{Break into chunks of size batch\_size }
	
	\hspace{0.5cm}\_batches = split(idx\_cases\_batch, batch\_size)
	
	\hspace{0.5cm}indexes = 1,...,|population| // \emph{An array of indexes}
	
	\hspace{0.5cm}
	
	\hspace{0.5cm}// \emph{Select k individuals}

	\hspace{0.5cm}selected\_individuals = {[} {]} // \emph{An empty collection}
	
	~	
    ~		
	\hspace{0.5cm}\textbf{while} |selected\_individuals| < k \textbf{do}
	
	\hspace{0.5cm}\hspace{0.5cm}batches = clone(\_batches) // \emph{Create the batch queue}
	
%	\hspace{0.5cm}\hspace{0.5cm}winners = {[} {]} // \emph{An empty collection}
	
	\hspace{0.5cm}
	
	\hspace{0.5cm}\hspace{0.5cm}\textbf{while} |batches| > 0 \textbf{and} |selected\_individuals|
	< k \textbf{do}
	
	\hspace{0.5cm}\hspace{0.5cm}\hspace{0.5cm}idx\_candidates = random\_selection(indexes, tourn\_size)
	// \emph{Tournament candidates per batch}
	
	~
	
	\hspace{0.5cm}\hspace{0.5cm}\hspace{0.5cm}// \emph{Calculate the candidate fitness for the
		batch}
	
	\hspace{0.5cm}\hspace{0.5cm}\hspace{0.5cm}cand\_fitness\_for\_this\_batch = {[} {]}
	
	\hspace{0.5cm}\hspace{0.5cm}\hspace{0.5cm}\textbf{for} idx \textbf{in} idx\_candidates:
	
	\hspace{0.5cm}\hspace{0.5cm}\hspace{0.5cm}\hspace{0.5cm}errors = {[} {]}
	
	\hspace{0.5cm}\hspace{0.5cm}\hspace{0.5cm}\hspace{0.5cm}\textbf{for} b \textbf{in} batches{[}1{]}:
	// \emph{Contains the index of the cases in the first batch in the
		queue}
	
	\hspace{0.5cm}\hspace{0.5cm}\hspace{0.5cm}\hspace{0.5cm}\hspace{0.5cm}errors.append (population{[}idx{]}.case\_error{[}idx\_cases\_batch{[}b{]]})
	
	\hspace{0.5cm}\hspace{0.5cm}\hspace{0.5cm}\hspace{0.5cm}\textbf{end for}
	
	\hspace{0.5cm}\hspace{0.5cm}\hspace{0.5cm}\hspace{0.5cm}cand\_fitness\_for\_this\_batch.append( mean(errors)
	)
	
	\hspace{0.5cm}\hspace{0.5cm}\hspace{0.5cm}\textbf{end for} 
	
	~
	
	\hspace{0.5cm}\hspace{0.5cm}\hspace{0.5cm}idx\_winner = index\_min(cand\_fitness\_for\_this\_batch)
	
	\hspace{0.5cm}\hspace{0.5cm}\hspace{0.5cm}winner = population{[}idx\_candidates{[}idx\_winner{]}{]}
	
	\hspace{0.5cm}\hspace{0.5cm}\hspace{0.5cm}selected\_individuals.append(winner) // \emph{Store the winner of each batch}
		
% 	\hspace{0.5cm}\hspace{0.5cm}\hspace{0.5cm}winners.append( population{[}idx\_candidates{[}idx\_winner{]}{]}
% 	)
	
	\hspace{0.5cm}\hspace{0.5cm}\hspace{0.5cm}batches.pop(1) // \emph{Remove the current batch
		from the queue}
	
	\hspace{0.5cm}\hspace{0.5cm}\textbf{end while}
	
% 	\hspace{0.5cm}\hspace{0.5cm}selected\_individuals.append\_each(winners) // \emph{Insert
% 		the winner of each batch}
	
	\hspace{0.5cm}\textbf{end while} 
	
	~
	
	\hspace{0.5cm}\textbf{return} selected\_individuals // \emph{Return a collection
		of individuals}
	
	\textbf{end function} 
	
	\end{flushleft}
	\label{bts_alg}	
	
\end{algorithm*}

\section{Experimental analysis}

\subsection{Computational environment}

All tests run in a machine with the following environment: Ubuntu 18.04, Anaconda Python 2.7.15, Deap 1.2.
The machine itself is an Intel(R) Xeon(R) Silver 4114 CPU @ 2.50GHz. 
The code is not numpy optimized. Numpy was used in
BTS, BTSS and Ae\_Lex only for calculating the mean, median, minimum,
shuffling and ordering the solutions.

\subsection{Data sets}

In the experiments, data sets that are widely employed in machine-learning
and symbolic regression \cite{oliveira2016dispersion, la2016epsilon} research were chosen.
%{[}Refs: Luis, Lacava{]}. 
As in \cite{la2016epsilon}, we normalized the features of the data sets (not the response) to zero mean, unit variance and randomly partitioned into
five disjoint sets of the same size. 
Later, we executed the methods five times (different seeds) with a 5-fold cross-validation (5 \texttimes{}
5-CV) for each configuration in a grid-search.

\begin{table}[!h]
	\caption{\label{tab:data sets}data sets and description.}
	
	\noindent \centering{}\resizebox{\columnwidth}{!}{%
		\begin{tabular}{cccc}
			\toprule 
			\textbf{\footnotesize{}Name} & \textbf{\footnotesize{}Variables} & \textbf{\footnotesize{}Training cases} & \textbf{\footnotesize{}Test cases}\tabularnewline
			\midrule
			{\footnotesize{}airfoil} & {\footnotesize{}5} & {\footnotesize{}1202} & {\footnotesize{}301}\tabularnewline
			{\footnotesize{}concrete} & {\footnotesize{}8} & {\footnotesize{}824} & {\footnotesize{}206}\tabularnewline
			{\footnotesize{}energyCooling} & {\footnotesize{}8} & {\footnotesize{}614} & {\footnotesize{}154}\tabularnewline
			{\footnotesize{}energyHeating} & {\footnotesize{}8} & {\footnotesize{}614} & {\footnotesize{}154}\tabularnewline
			{\footnotesize{}towerData} & {\footnotesize{}25} & {\footnotesize{}3999} & {\footnotesize{}1000}\tabularnewline
			{\footnotesize{}wineRed} & {\footnotesize{}11} & {\footnotesize{}1279} & {\footnotesize{}320}\tabularnewline
			{\footnotesize{}wineWhite} & {\footnotesize{}11} & {\footnotesize{}3918} & {\footnotesize{}980}\tabularnewline
			{\footnotesize{}yacht} & {\footnotesize{}6} & {\footnotesize{}614} & {\footnotesize{}154}\tabularnewline
			\bottomrule
	\end{tabular}}
\end{table}

\subsection{GP configuration}

The general parameter settings for the algorithms are shown
in Table\ \ref{tab:General-parameter-settings.} and follows the configuration used in \cite{la2016epsilon}. 
MAE was chosen as the fitness function to guide the evolution process.
Moreover, we performed a grid-search on batch and tournament sizes for both BTS and the canonical tournament.

\begin{table}[!htb]
	\caption{\label{tab:General-parameter-settings.}General parameter settings.
		Operators are protected, meaning they return 1.0 on failure.}
	
	\noindent \centering{}\resizebox{\columnwidth}{!}{%
		\begin{tabular}{cc}
			\toprule 
			\textbf{\footnotesize{}Parameter} & \textbf{\footnotesize{}Setting}\tabularnewline
			\midrule
			{\footnotesize{}Independent runs} & {\footnotesize{}25}\tabularnewline
			{\footnotesize{}Initialization} & {\footnotesize{}Ramped half/half}\tabularnewline
			{\footnotesize{}Number of generations} & {\footnotesize{}1000}\tabularnewline
			{\footnotesize{}Population size} & {\footnotesize{}1000}\tabularnewline
			{\footnotesize{}Maximum initial depth} & {\footnotesize{}3}\tabularnewline
			{\footnotesize{}Maximum depth} & {\footnotesize{}7}\tabularnewline
			{\footnotesize{}Batch sizes} & {\footnotesize{}2, 4, 8, 16, 32, 64, 128}\tabularnewline
			{\footnotesize{}Tournament sizes} & {\footnotesize{}2, 4, 8, 16, 32, 64, 128}\tabularnewline
			{\footnotesize{}Shuffle} & {\footnotesize{}True, False}\tabularnewline
			{\footnotesize{}Elite} & {\footnotesize{}1}\tabularnewline
			{\footnotesize{}Crossover operator} & {\footnotesize{}One-point}\tabularnewline			
			{\footnotesize{}Crossover rate} & {\footnotesize{}90\%}\tabularnewline
			{\footnotesize{}Mutation operator} & {\footnotesize{}Uniform (new random subtree)}
			 \tabularnewline
			{\footnotesize{}Mutation rate} & {\footnotesize{}10\%}\tabularnewline						 
			{\footnotesize{}Terminal set} & {\footnotesize{}\{x, ERC, +, \textminus , \textasteriskcentered ,
				/, sin, cos, exp, log\}}\tabularnewline
			{\footnotesize{}ERC range} & {\footnotesize{}[-1, 1]}\tabularnewline
			\bottomrule
	\end{tabular}}
\end{table}

\subsection{Analysis}

% In this paper, we analyzed how our approach performs in many configurations and compared each of them, individually, to Ae-Lex. The main question we intend to answer is: can BTS or BTSS be competitive to Ae-Lex in terms of solution quality but with a lower computational cost? Hence, the objective is not to find the perfect configuration since there is no free lunch.

We analyzed how our approaches perform in many configurations and later compared each of them, individually, to Ae-Lex. The objective is to investigate how the parameters affect the search and to evaluate the general performance against Ae-Lex. Therefore, the key question we are trying to answer is: can BTS or BTSS be competitive to Ae-Lex in terms of solution quality but considerably faster? 

We do not intend to find the perfect configuration that is the best for all problems since there is no free lunch. In fact, we intend to observe, from the experiments, which configurations perform best for the specific data sets and GP configuration used in this paper that can be suggested as initial trials for GP practitioners and which do not show a good performance and could be avoided.

To reduce the configurations for comparison, for each data set, we first select the top five configurations according
to the Median MAE (MMAE) on the \emph{training set}. 
Then, we summarize the experimental results on the \emph{test set} (Note that we are not selecting the best on the test set).
To analyze the behavior of the proposed method as well as compare it with existing ones, the median performance curve of the best fitness (calculated
as MAE) over all runs, a boxplot of the best fitness and running time are used.

Moreover, we present a hypothesis test using Wilcoxon Rank Sum
with Holm correction.
In this context, $\alpha=5\%$ is used to evaluate how the methods performed against Ae-Lex (baseline) on each data set.

% Later, we present a hypothesis test using pairwise-Wilcoxon Rank Sum
% with Holm correction. $\alpha=5\%$ to evaluate how the methods performed
% against each other on each data set.

\subsection{Results and discussion}

Here we will adopt the following nomenclature: 1) BTS/$bs$/$ts$, 2) BTSS/$bs$/$ts$, and 3) Tourn/$ts$ in which Tourn refers to tournament selection, BTS and BTSS are the proposed algorithms and both $bs$ and $ts$ stands for respectively the batch size and tournament size. 
Notice that BTS is batch tournament selection in which cases are ordered by their
difficulty, i.e., differently from BTSS no shuffling is employed. 
Moreover, we will use Ae-Lex to mean Automatic $\epsilon$-Lexicase
selection.

\subsubsection{Performance Analysis}
\begin{figure*}[!t]
	\noindent \begin{centering}
		\begin{tabular}{cccc}
			\includegraphics[width=0.23\textwidth]{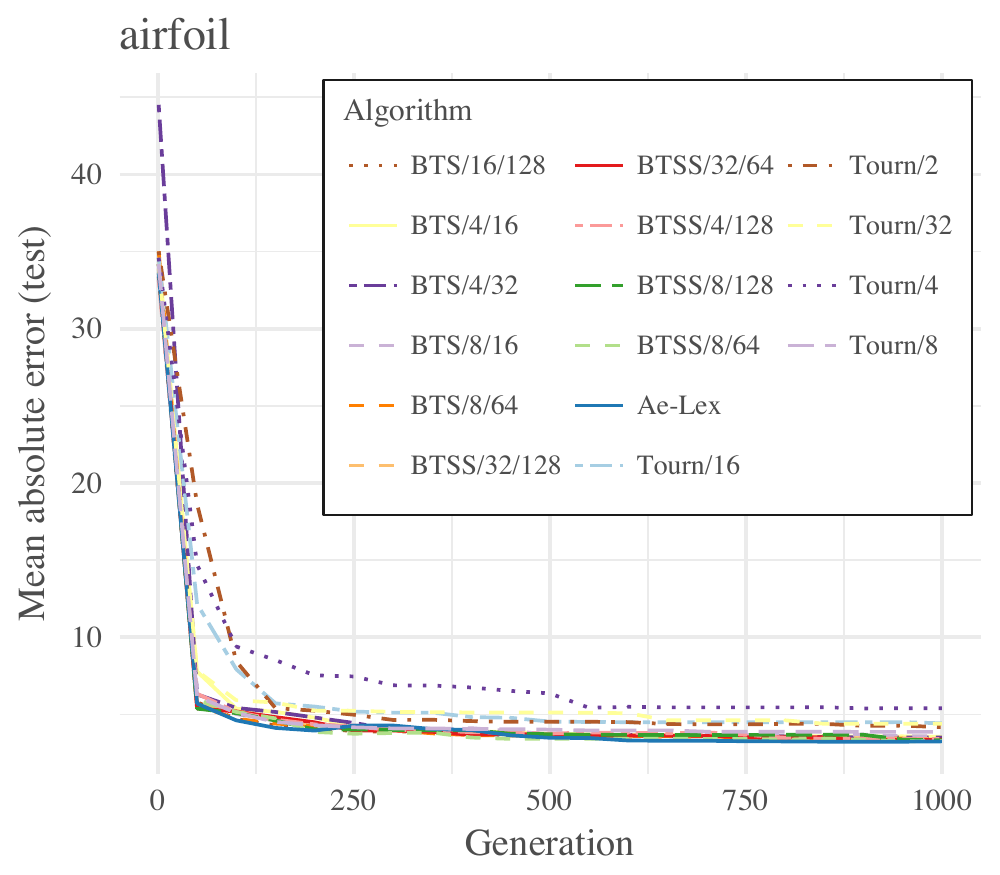} & \includegraphics[width=0.23\textwidth]{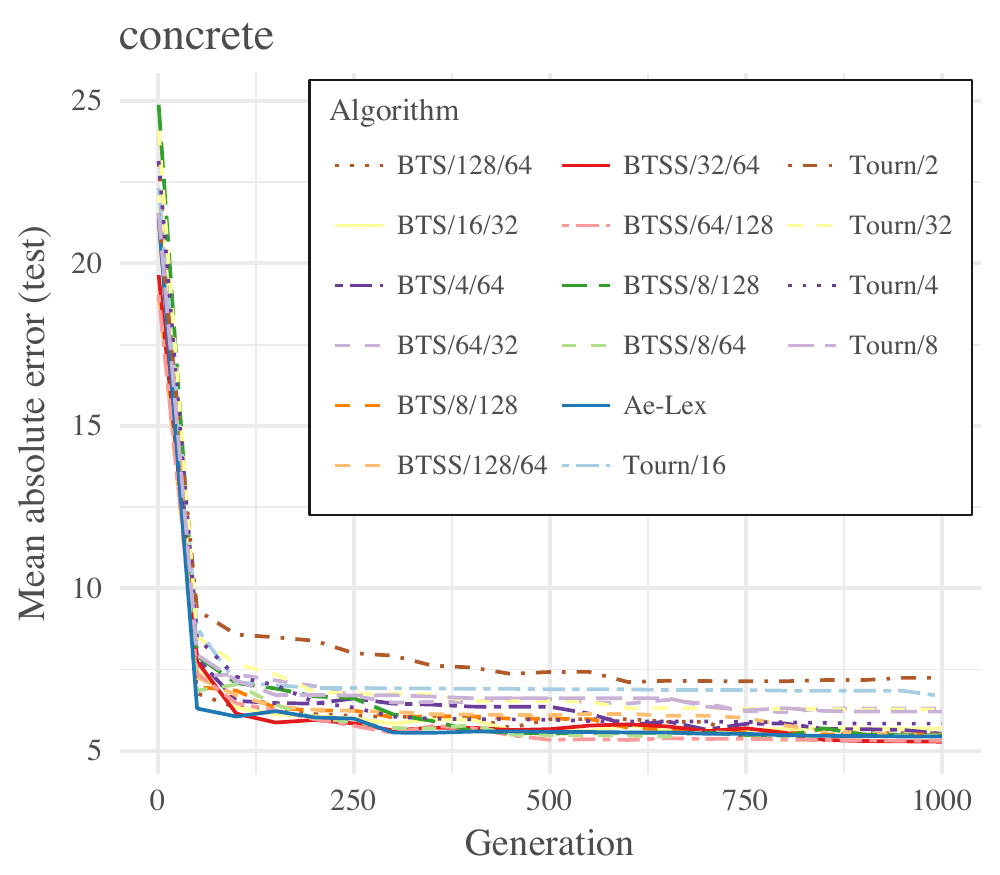} & \includegraphics[width=0.23\textwidth]{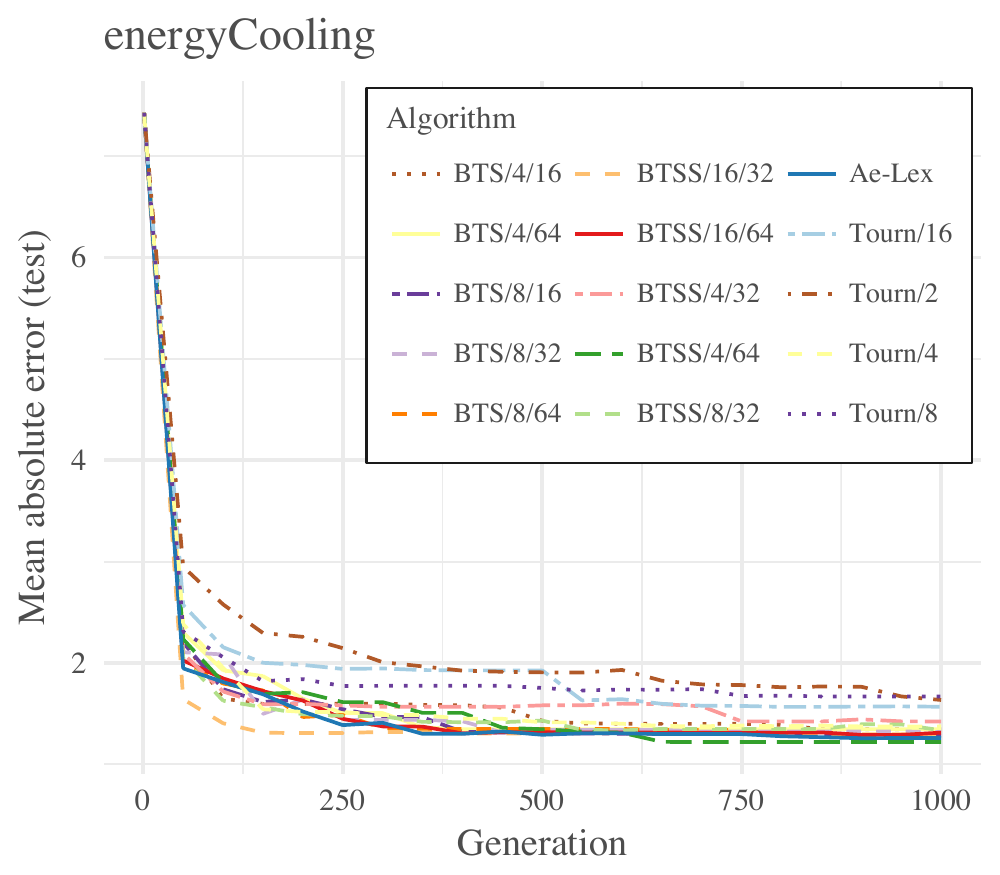} & \includegraphics[width=0.23\textwidth]{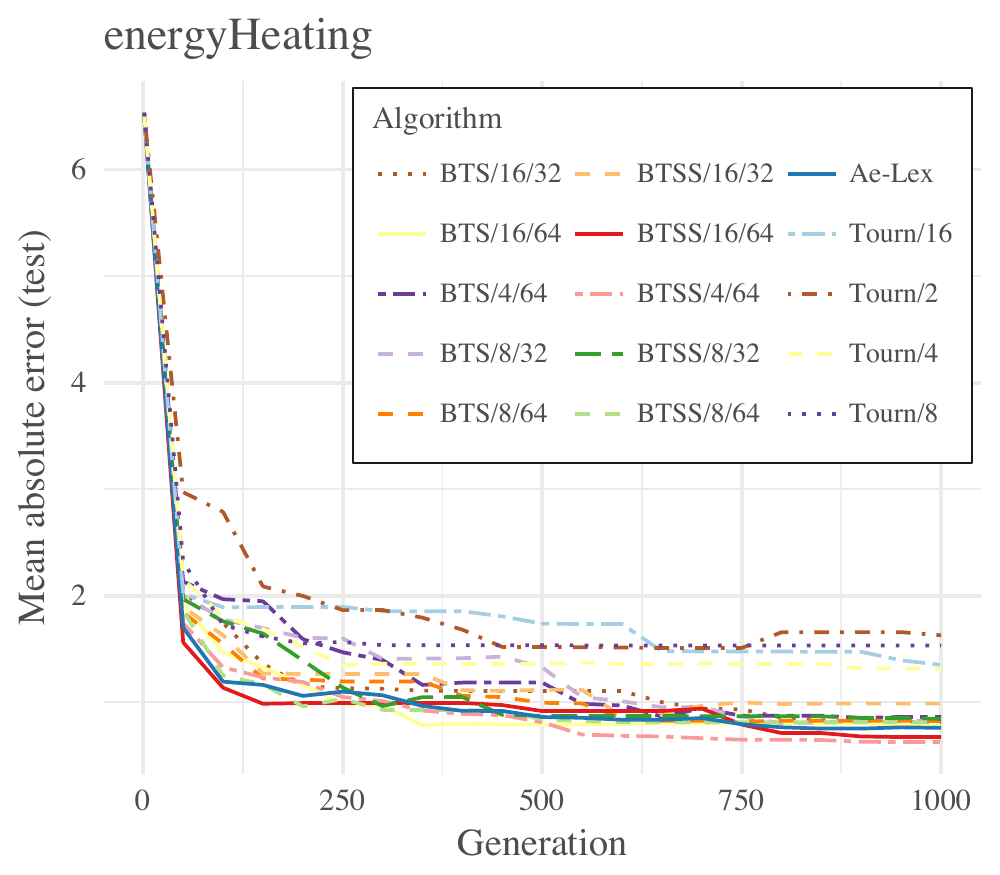}\tabularnewline
			\includegraphics[width=0.23\textwidth]{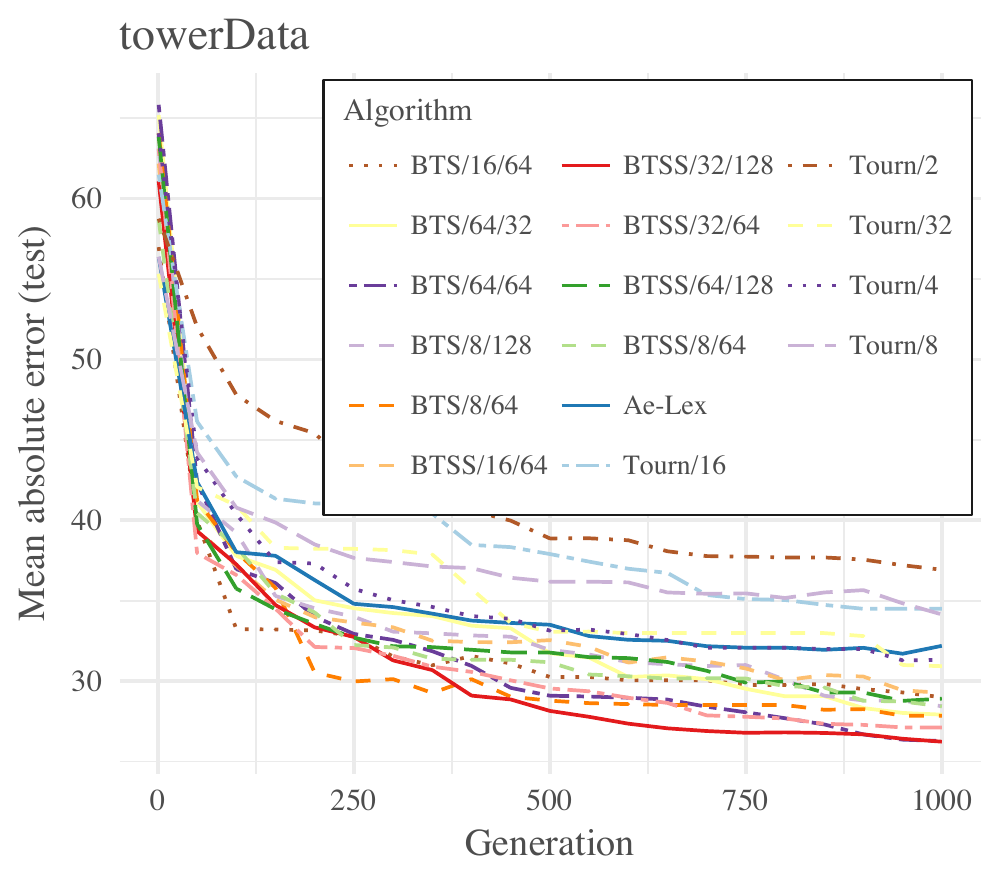} & \includegraphics[width=0.23\textwidth]{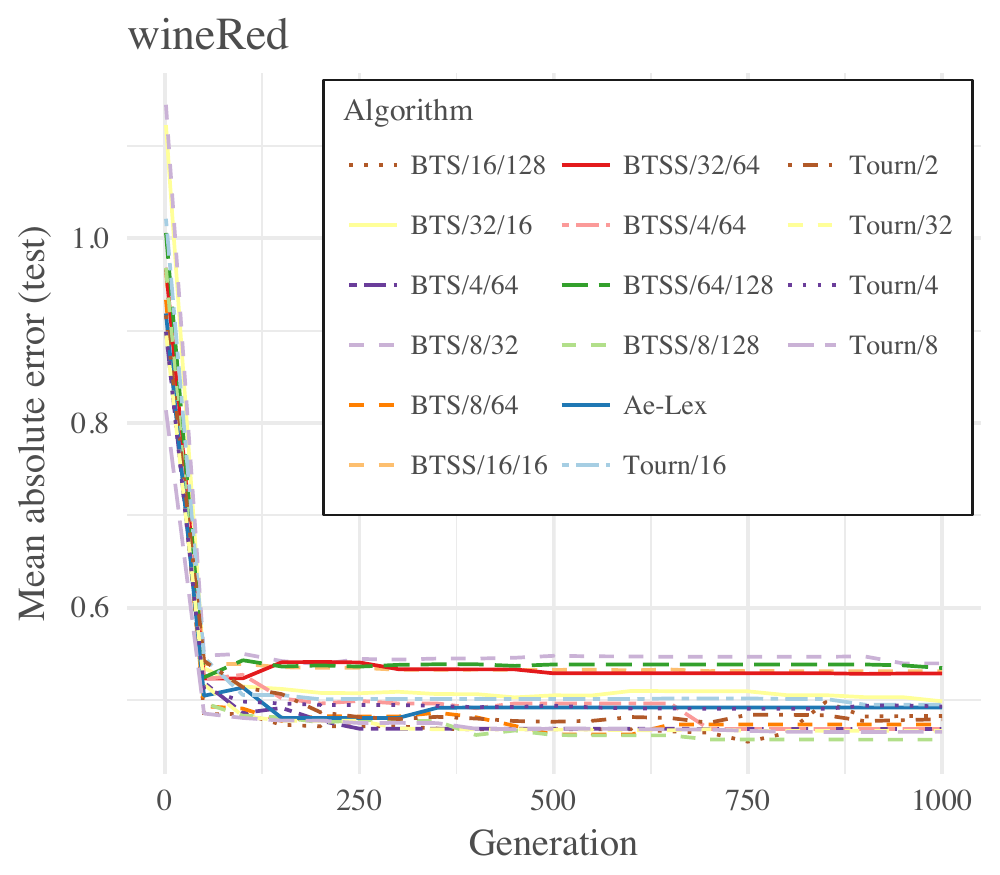} & \includegraphics[width=0.23\textwidth]{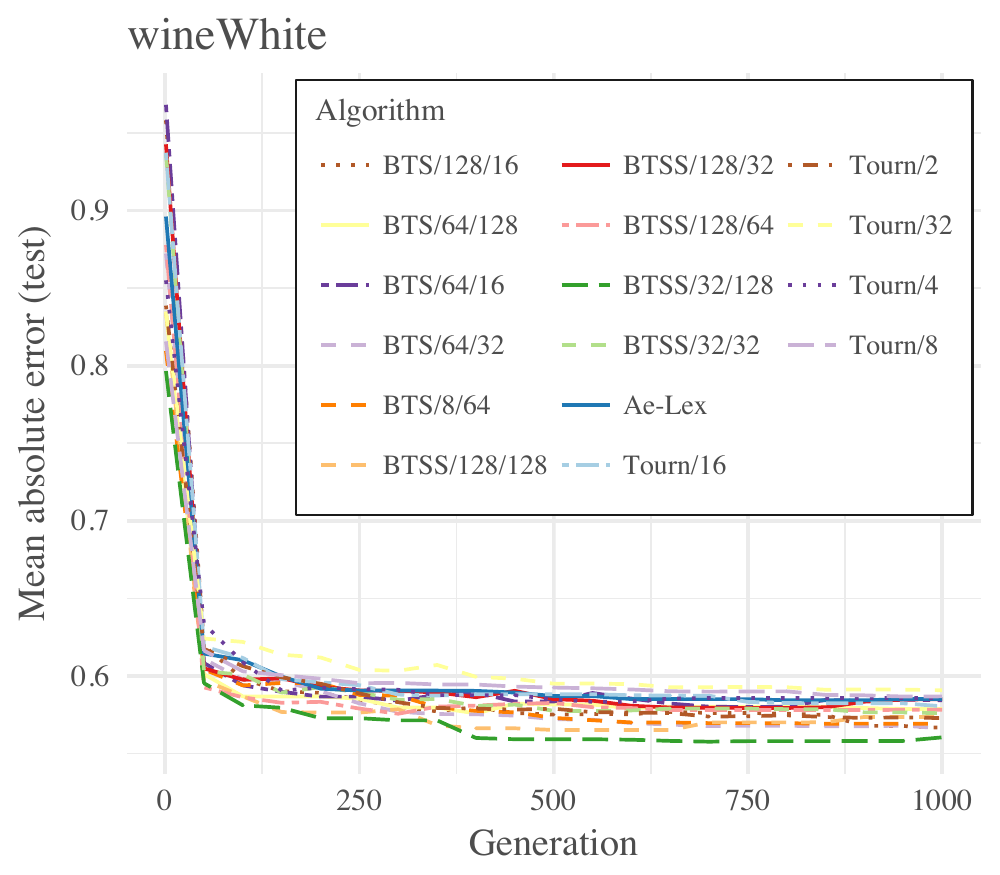} & \includegraphics[width=0.23\textwidth]{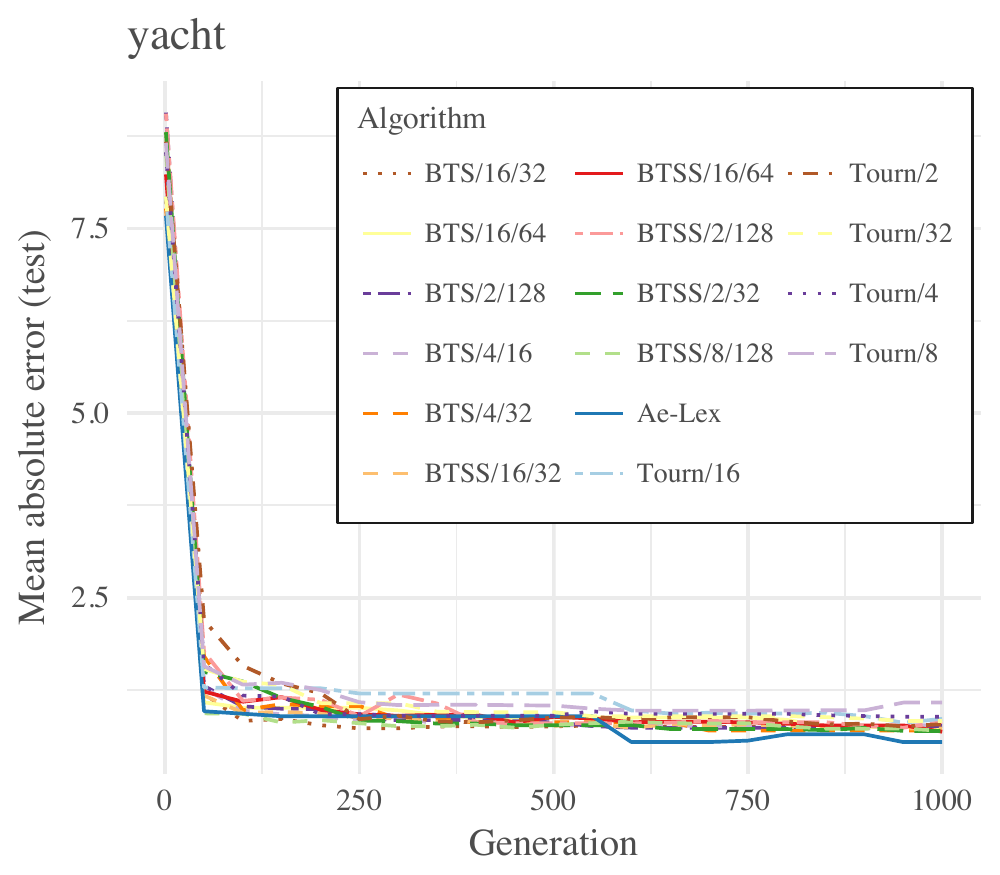}\tabularnewline
		\end{tabular}
		\par\end{centering}
	\caption{\label{fig:Curves-mae}Curves showing the MMAE on the test set.}
\end{figure*}

Figure~\ref{fig:Curves-mae} shows the performance curves for the top five configurations on each data set. 
BTS and BTSS perform similarly to Ae\_Lex while the canonical tournament selection shows the worst overall performance.  
On towerData, wineRed, and wineWhite data sets, BTS/BTSS are shown to easily outperform Ae\_Lex and tournament while achieving a similar performance  to Ae\_Lex on the other data sets.
Tournament selection shows the highest MAE on most data sets, with the worst performance on towerData data set. 

\begin{figure*}[!t]
	\noindent \begin{centering}
		\begin{tabular}{cccc}
			\includegraphics[width=0.23\textwidth]{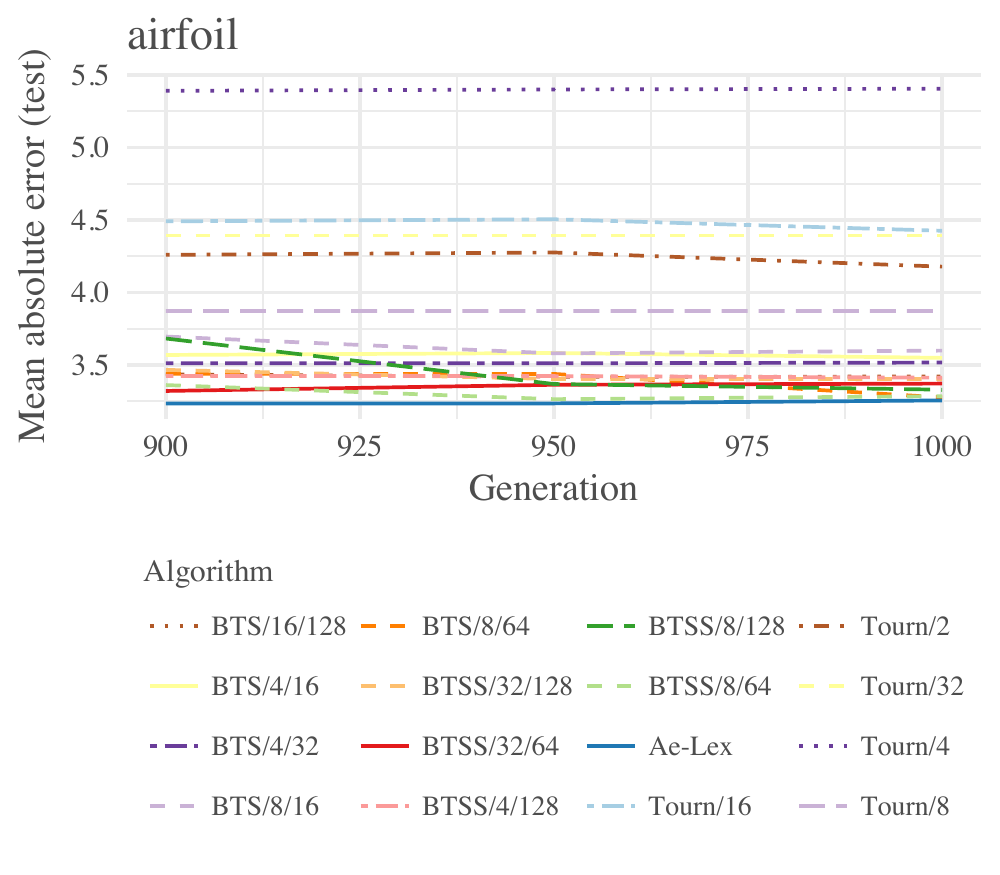} & \includegraphics[width=0.23\textwidth]{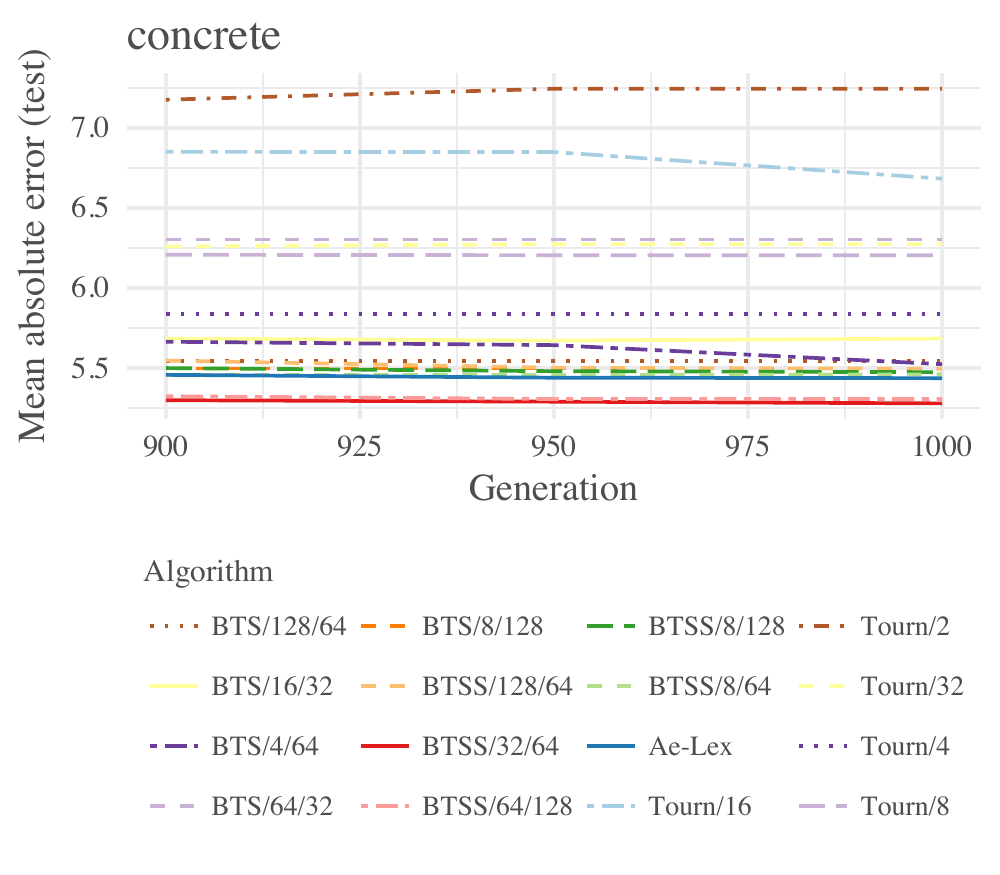} & \includegraphics[width=0.23\textwidth]{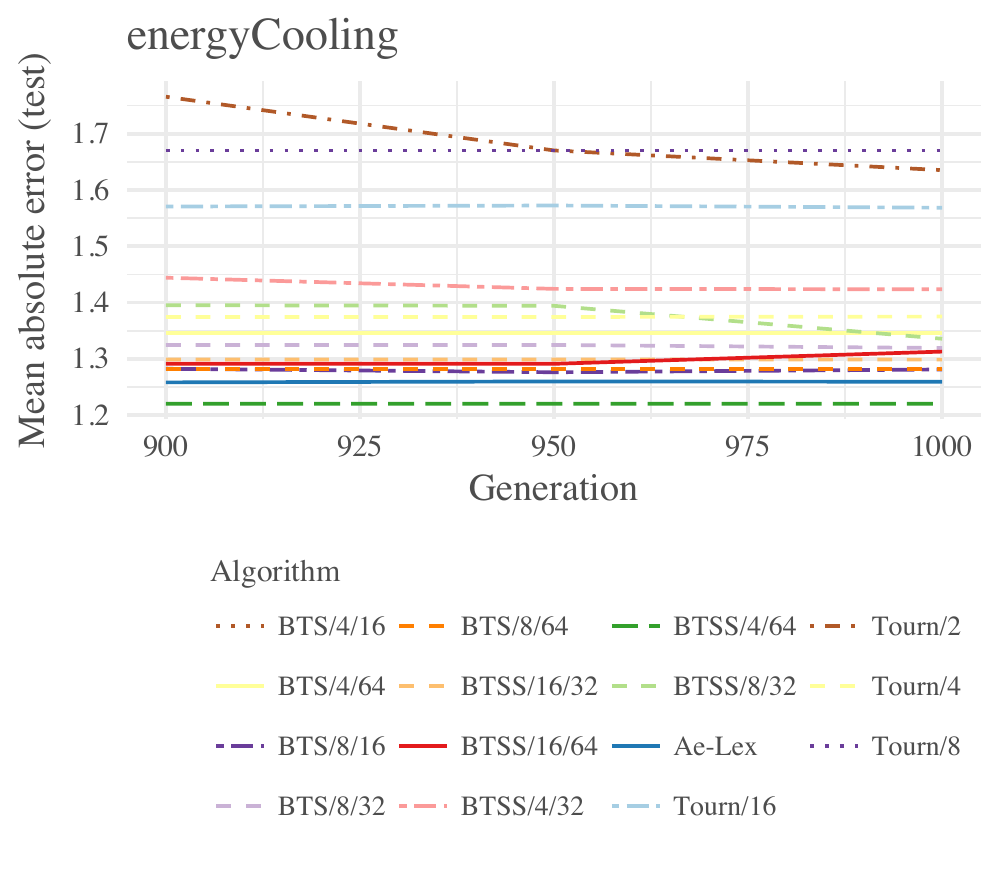} & \includegraphics[width=0.23\textwidth]{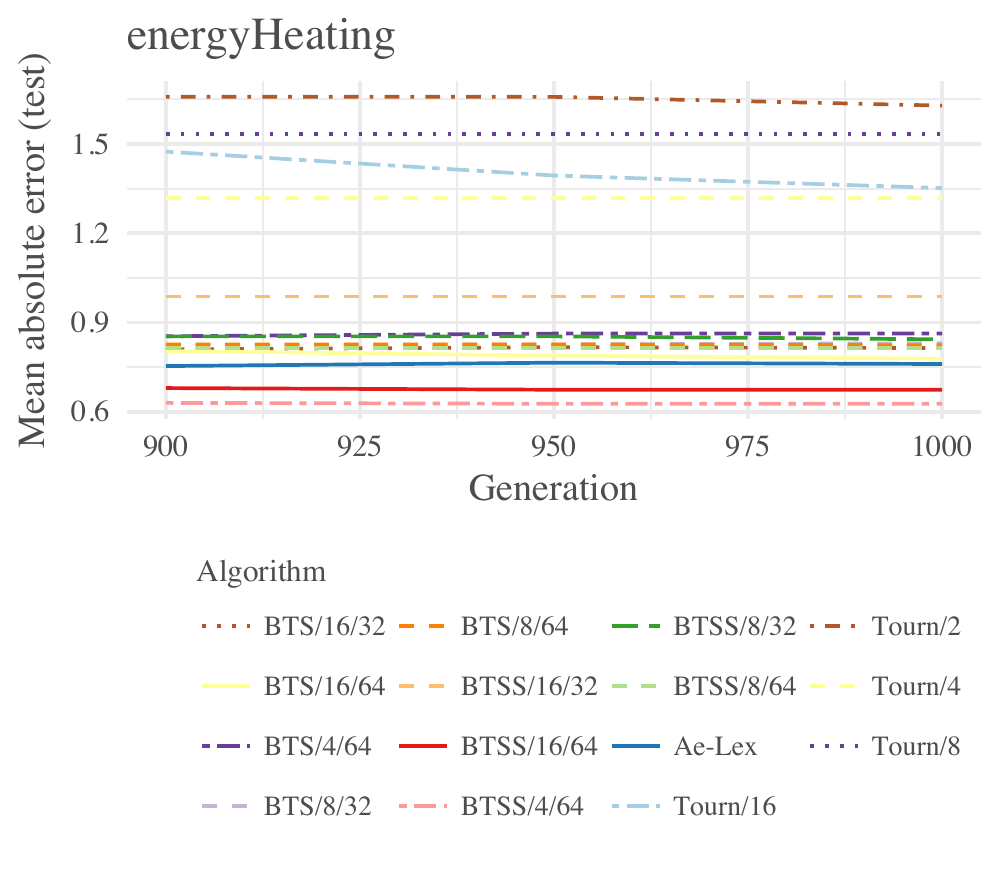}\tabularnewline
			\includegraphics[width=0.23\textwidth]{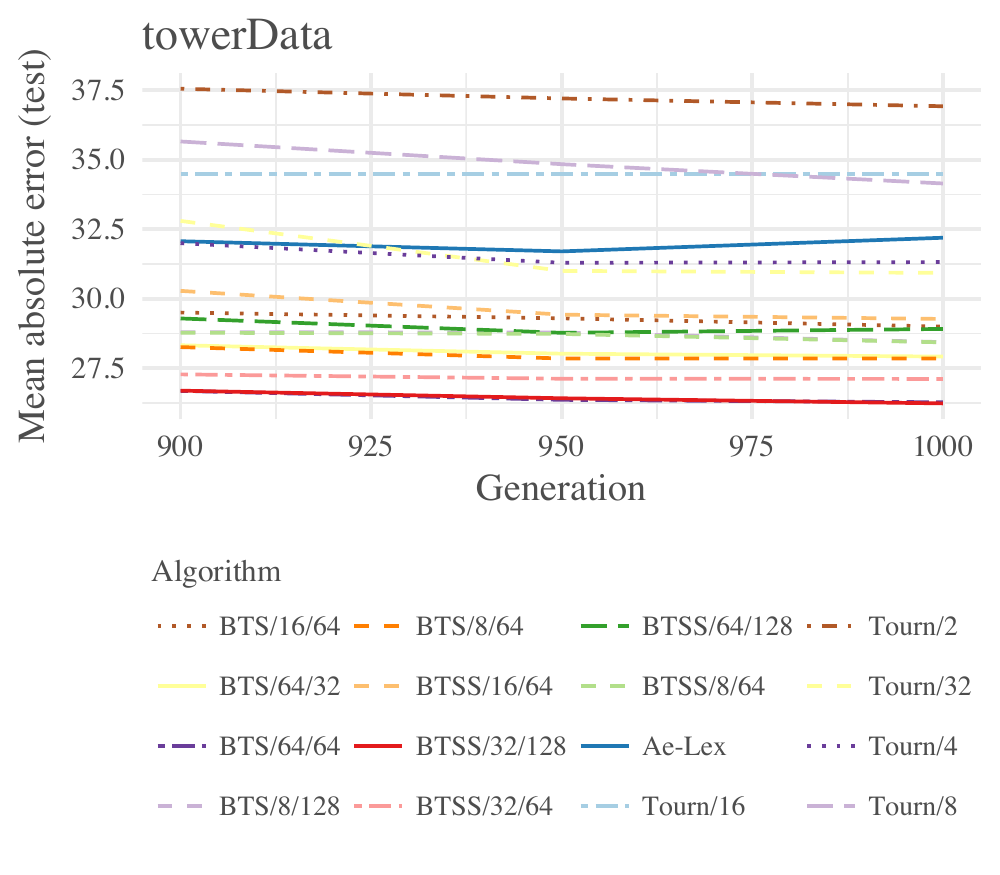} & \includegraphics[width=0.23\textwidth]{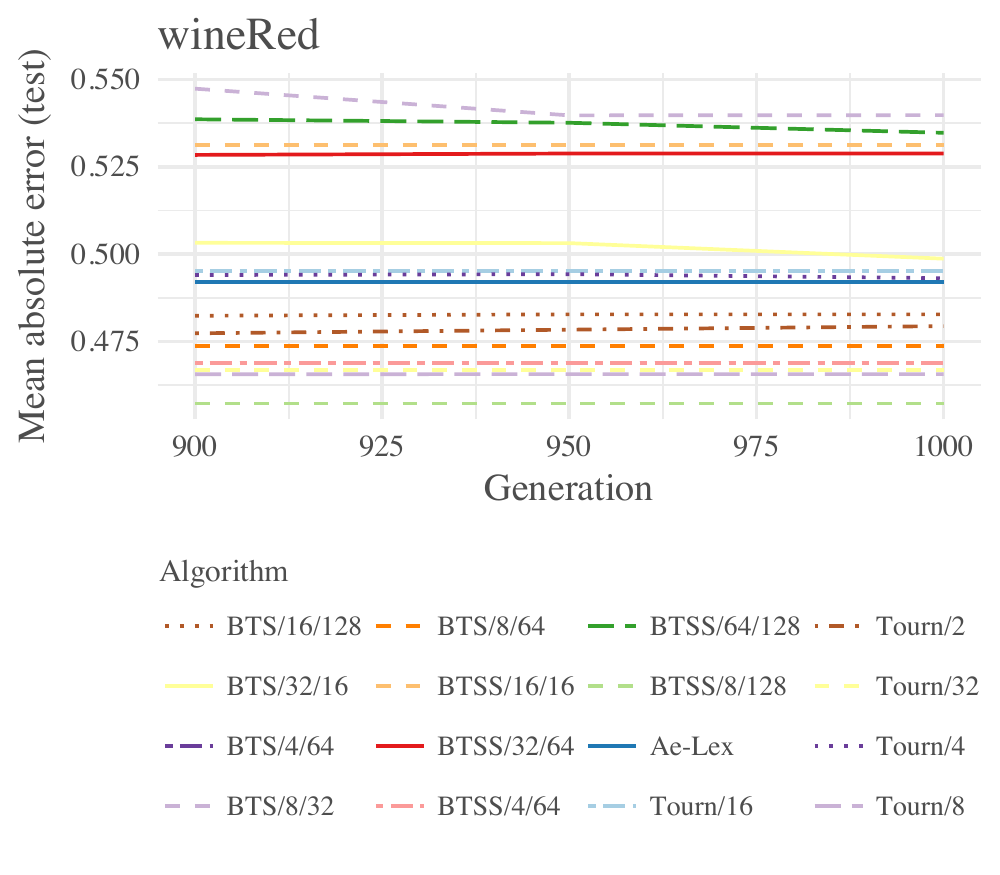} & \includegraphics[width=0.23\textwidth]{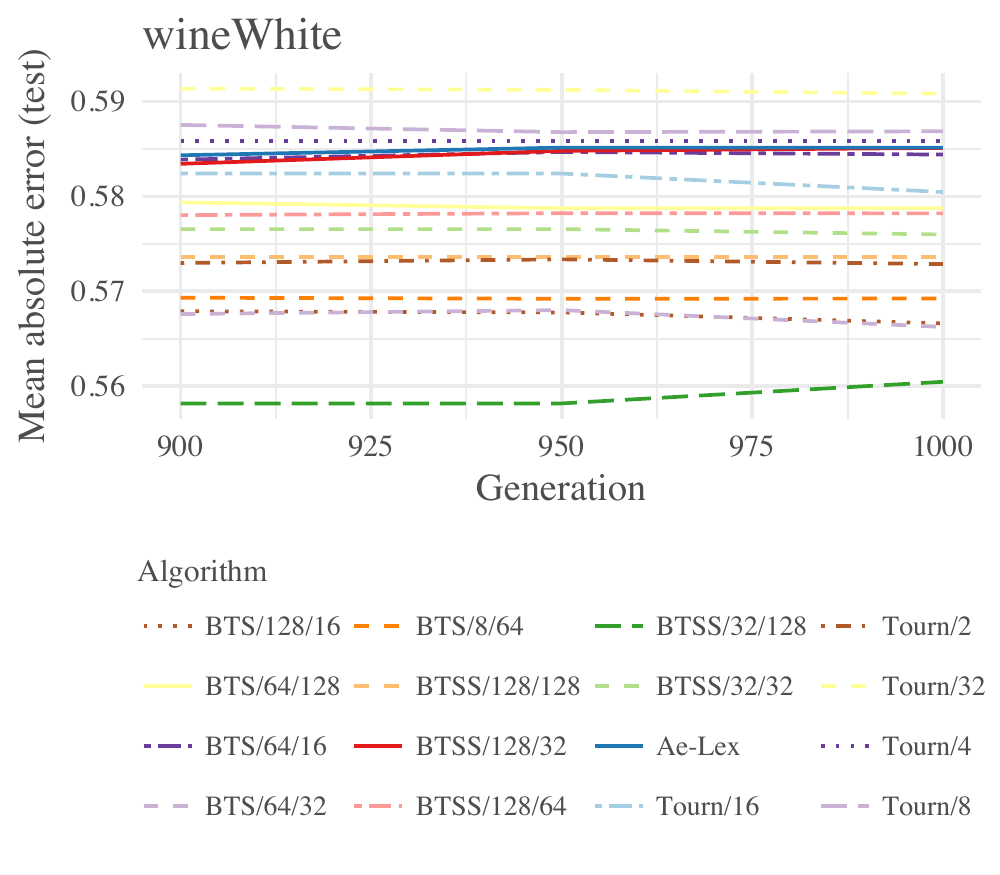} & \includegraphics[width=0.23\textwidth]{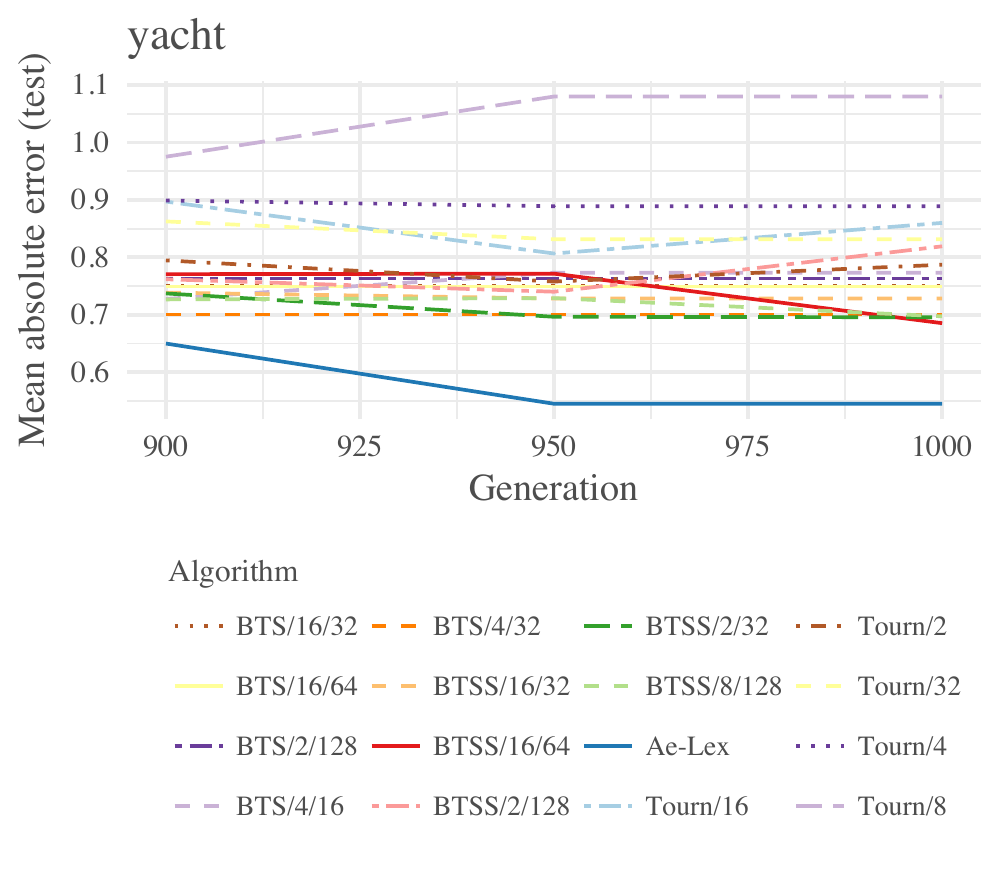}\tabularnewline
		\end{tabular}
		\par\end{centering}
	\caption{\label{fig:Curves-mae-zoom}Curves zoomed on the last 100 generations
		showing the MMAE on the test set.}
\end{figure*}

\begin{figure*}[!t]
	\noindent \begin{centering}
		\begin{tabular}{cccc}
			\includegraphics[width=0.23\textwidth]{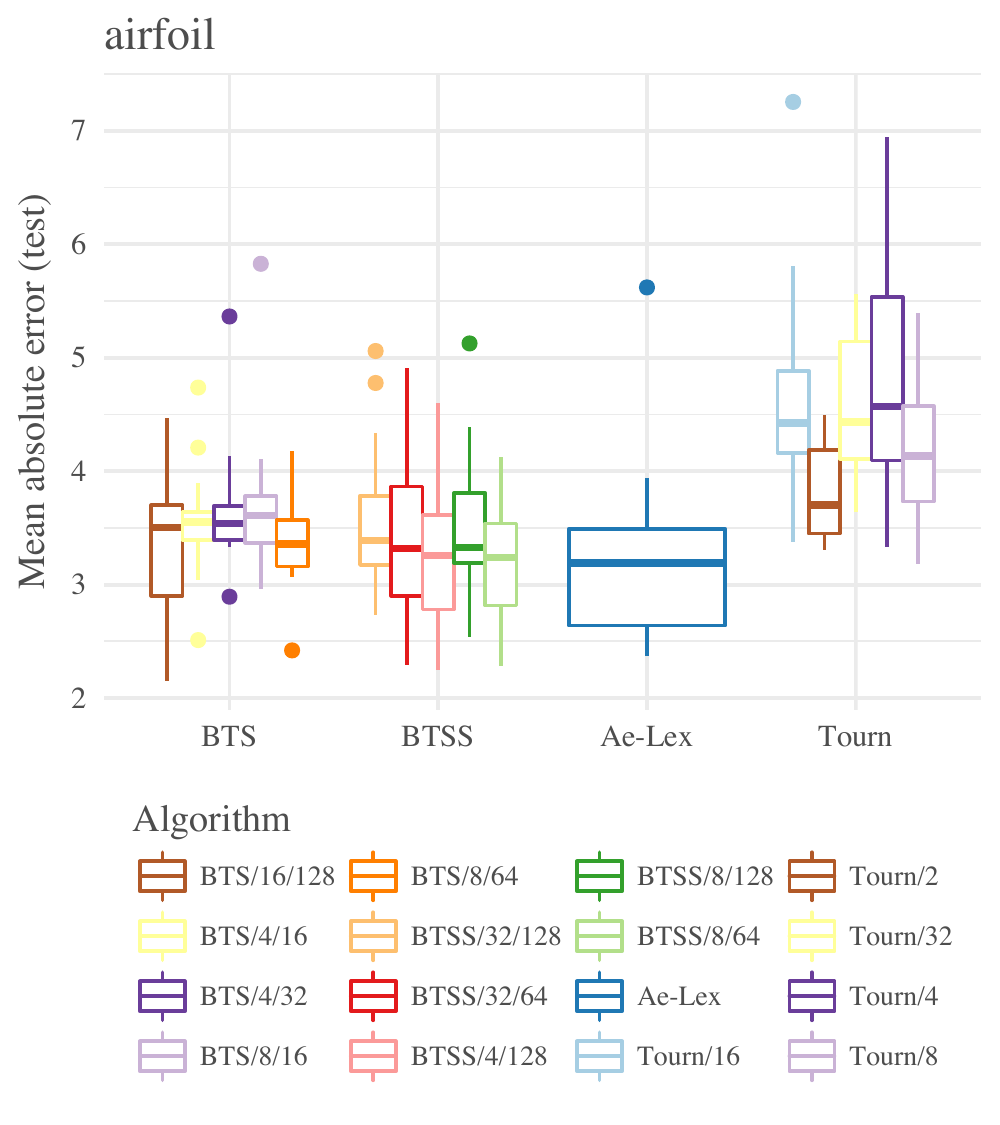} & \includegraphics[width=0.23\textwidth]{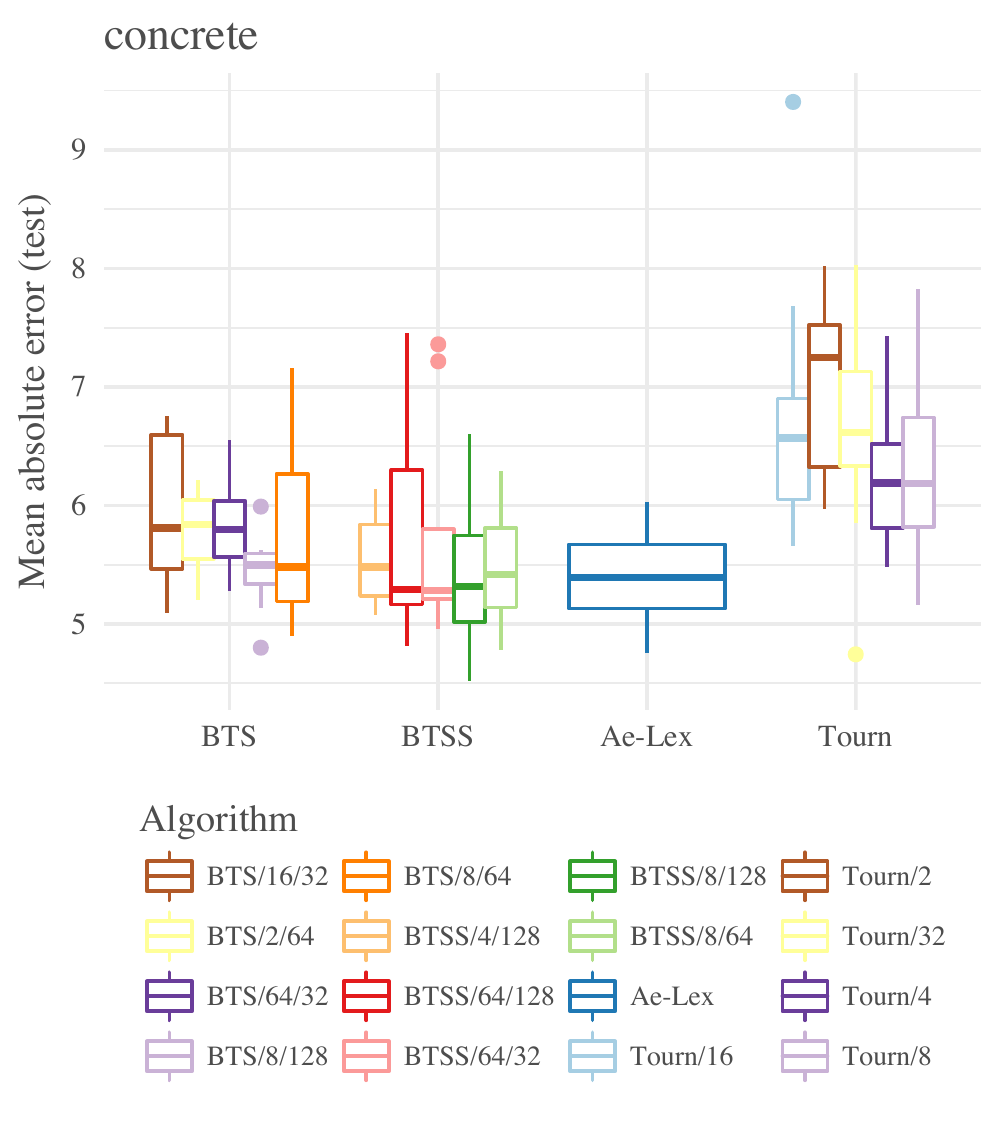} & \includegraphics[width=0.23\textwidth]{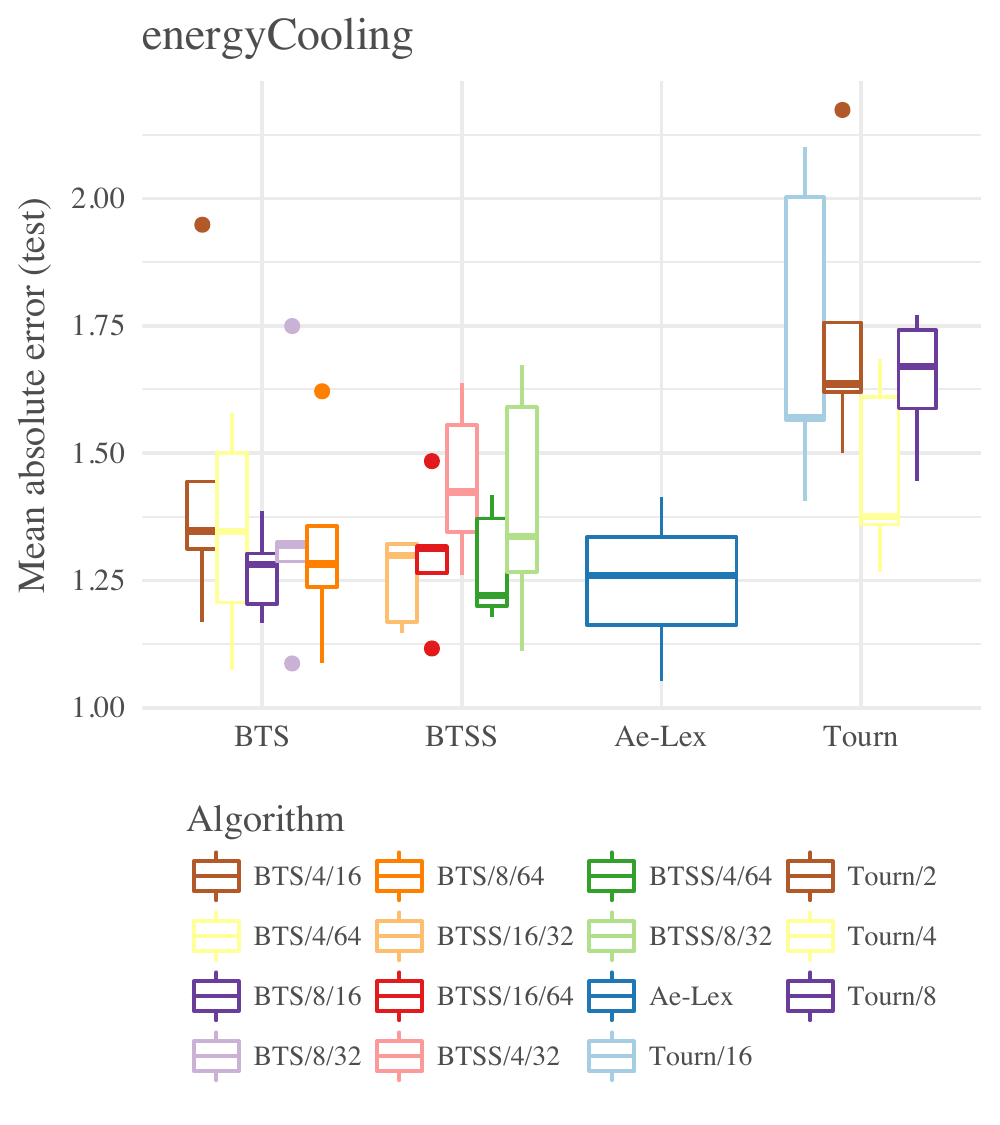} & \includegraphics[width=0.23\textwidth]{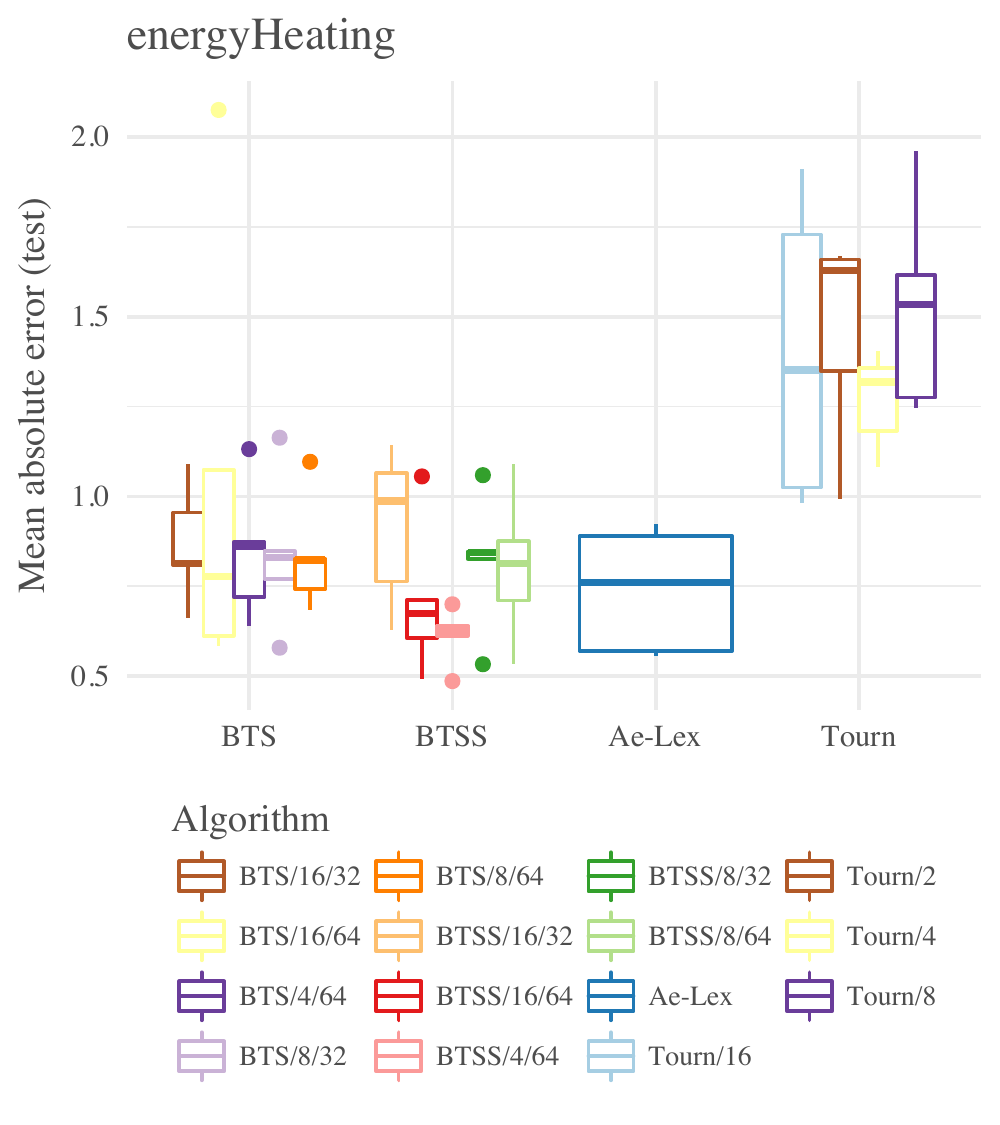}\tabularnewline
			\includegraphics[width=0.23\textwidth]{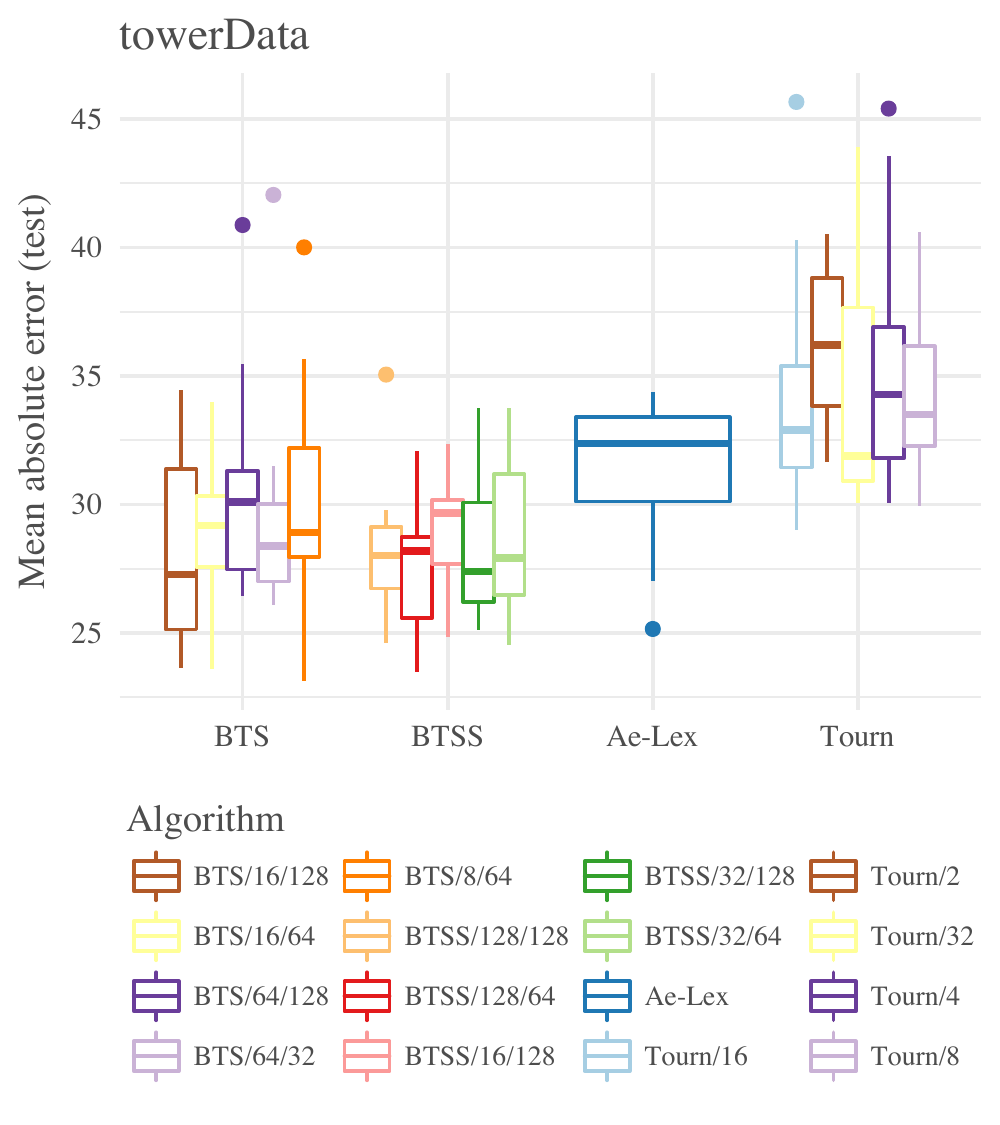} & \includegraphics[width=0.23\textwidth]{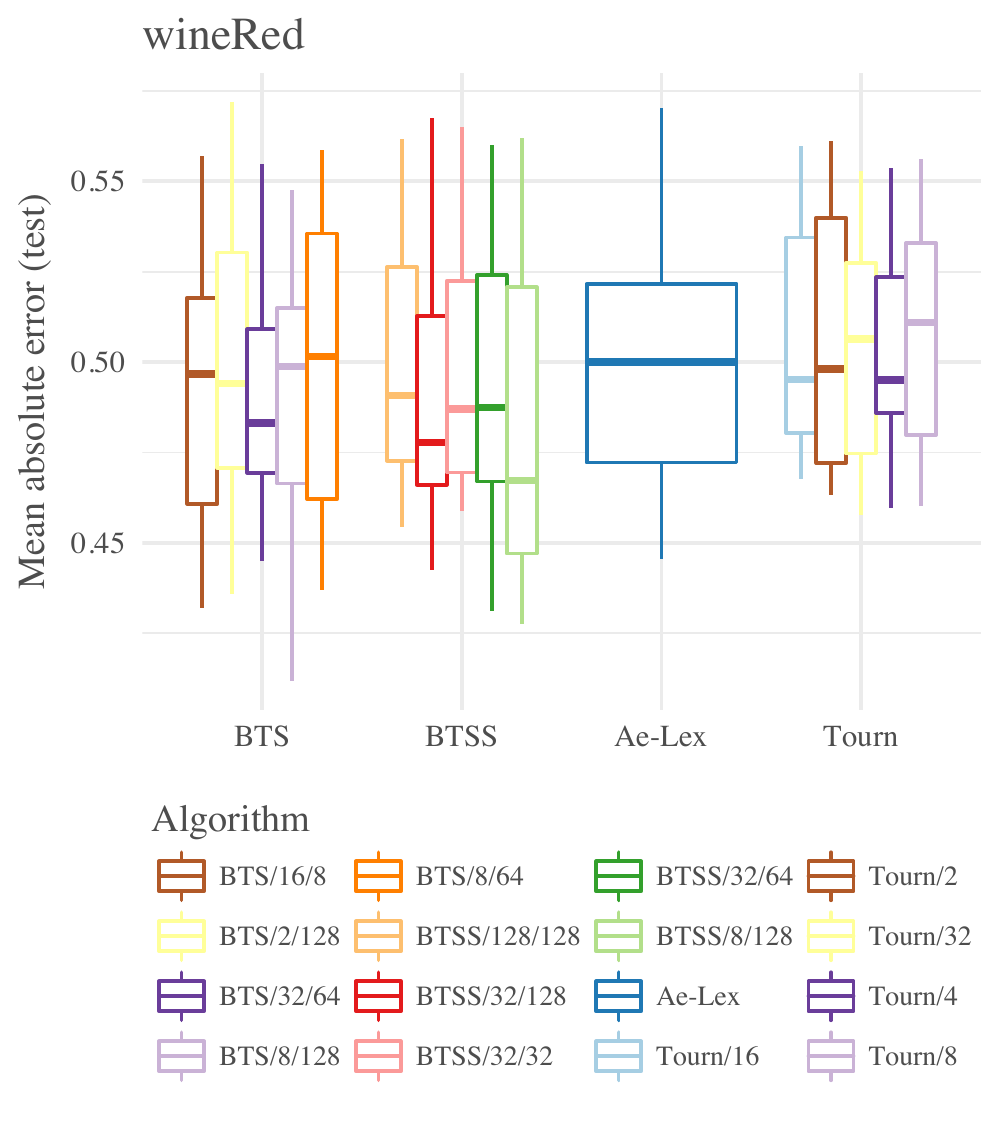} & \includegraphics[width=0.23\textwidth]{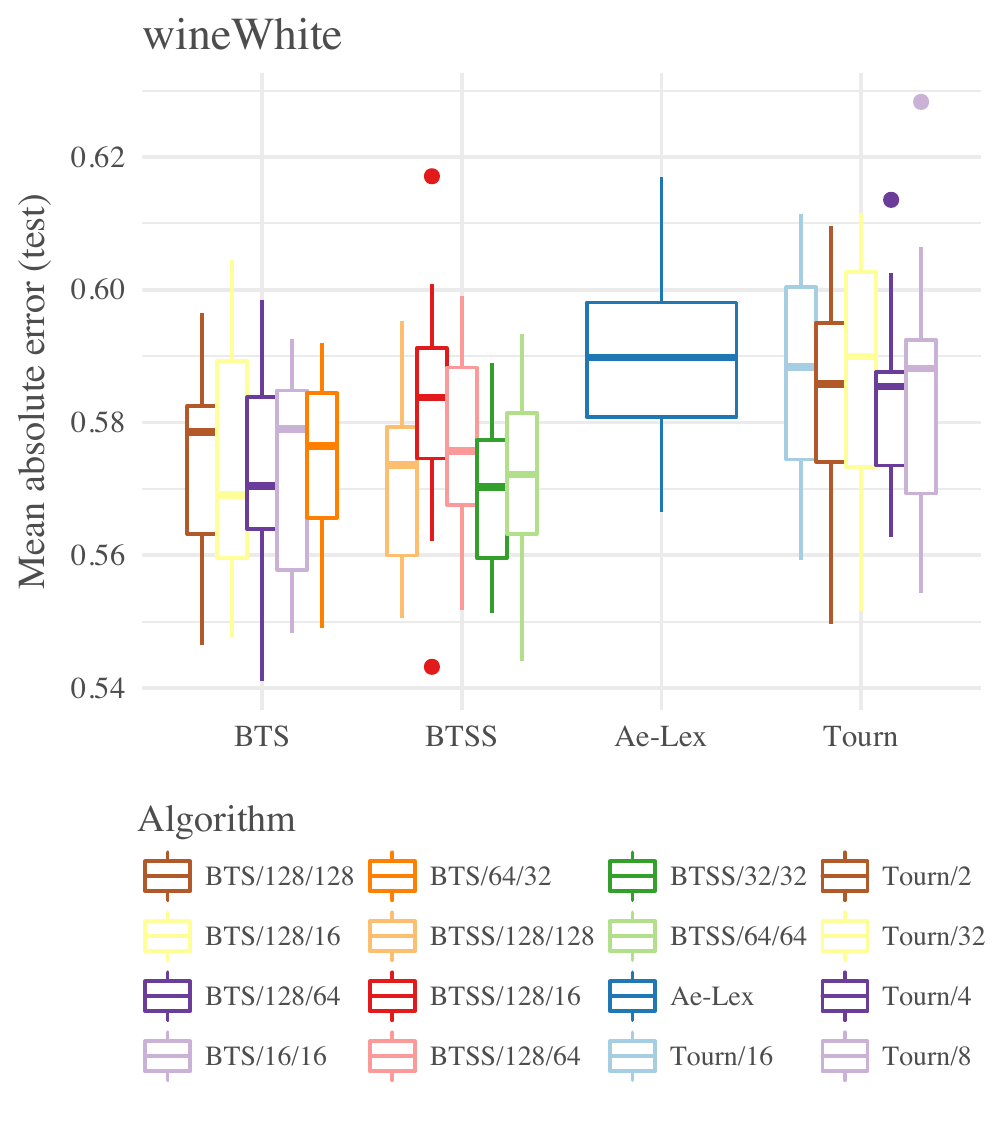} & \includegraphics[width=0.23\textwidth]{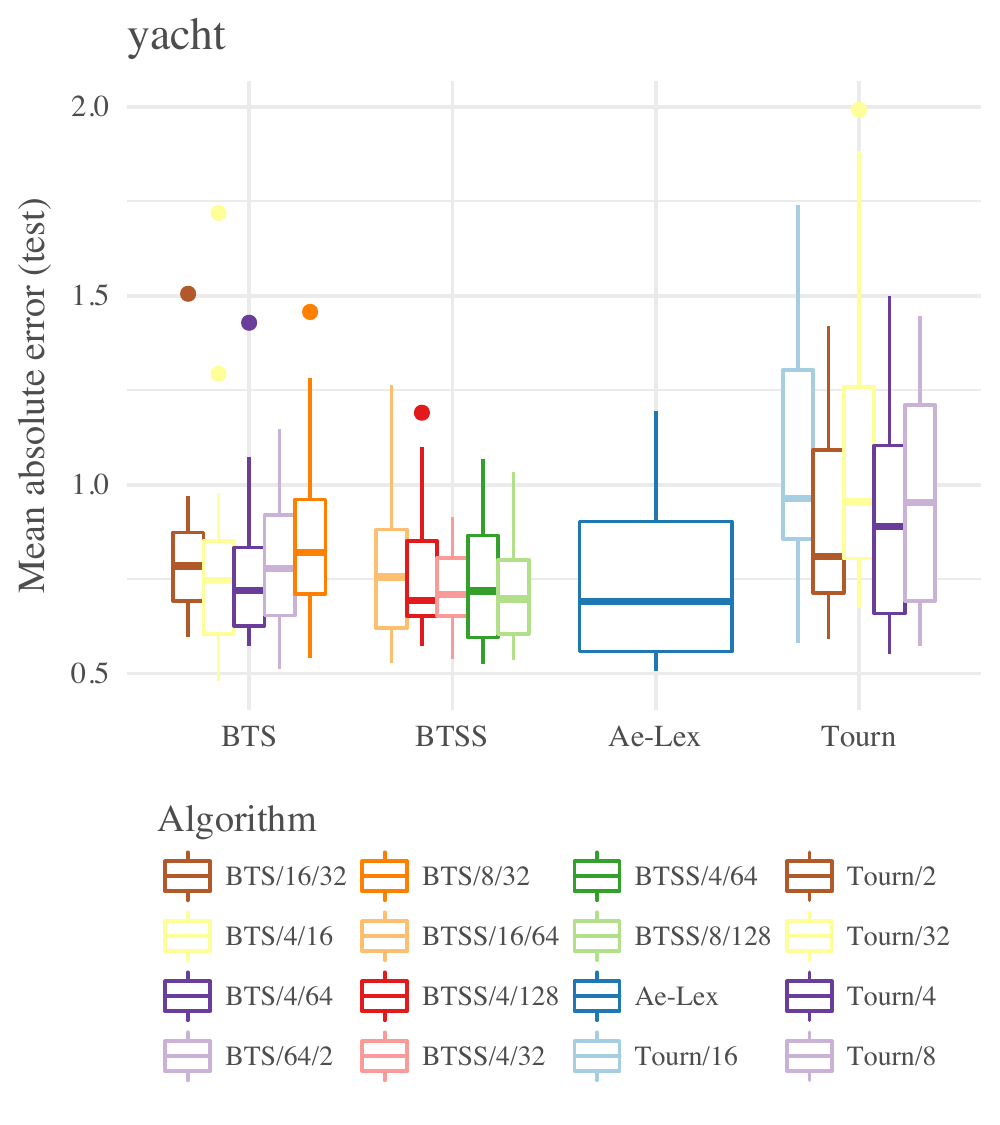}\tabularnewline
		\end{tabular}
		\par\end{centering}
	\caption{\label{fig:Boxplots-mae}Boxplots of the MAE on the test set.}
\end{figure*}

However, many of the curves overlap in the last generations.
Aside from towerData, wineRed and whiteRed, it is hard to tell if any algorithms had a superior performance.
To solve this problem and allow for a better analysis, in Figure~\ref{fig:Curves-mae-zoom} the last 100 generations are plotted separately. 
Now it is possible to observe that BTS/BTSS had similar or better performance than Ae-Lex in most of the data sets with the only exception being the yacht one.
MAE related boxplots (Figure~\ref{fig:Boxplots-mae}) further confim the results observed in Figures~\ref{fig:Curves-mae} and~\ref{fig:Curves-mae-zoom}.

%Ae\_Lex had the best median curve for airfoil and yacht data sets, but failed on the toiwer, wineRed, and wineWhite. For the other data sets, Ae\_Lex was close to the best.
%\begin{figure*}[!t]
%	\noindent \begin{centering}
%		\begin{tabular}{ccc}
%			\includegraphics[width=0.32\textwidth]{images/curves_test_mae_airfoil_} & \includegraphics[width=0.32\textwidth]{images/curves_test_mae_concrete_} & \includegraphics[width=0.32\textwidth]{images/curves_test_mae_energyCooling_} \tabularnewline
%			\includegraphics[width=0.32\textwidth]{images/curves_test_mae_energyHeating_} & \includegraphics[width=0.32\textwidth]{images/curves_test_mae_towerData_} & \includegraphics[width=0.32\textwidth]{images/curves_test_mae_wineRed_} \tabularnewline
%			\includegraphics[width=0.32\textwidth]{images/curves_test_mae_wineWhite_} & \includegraphics[width=0.32\textwidth]{images/curves_test_mae_yacht_}\tabularnewline
%		\end{tabular}
%		\par\end{centering}
%	\caption{\label{fig:Curves-mae}Curves showing the MMAE on the test set.}
%\end{figure*}

%\begin{figure*}[!t]
%	\noindent \begin{centering}
%		\begin{tabular}{ccc}
%			\includegraphics[width=0.32\textwidth]{images/curveszoom_test_mae_airfoil_} & \includegraphics[width=0.32\textwidth]{images/curveszoom_test_mae_concrete_} & \includegraphics[width=0.32\textwidth]{images/curveszoom_test_mae_energyCooling_} \tabularnewline
%			\includegraphics[width=0.32\textwidth]{images/curveszoom_test_mae_energyHeating_} & \includegraphics[width=0.32\textwidth]{images/curveszoom_test_mae_towerData_} & \includegraphics[width=0.32\textwidth]{images/curveszoom_test_mae_wineRed_} \tabularnewline
%			\includegraphics[width=0.32\textwidth]{images/curveszoom_test_mae_wineWhite_} & \includegraphics[width=0.32\textwidth]{images/curveszoom_test_mae_yacht_}\tabularnewline
%		\end{tabular}
%		\par\end{centering}
%	\caption{\label{fig:Curves-mae-zoom}Curves zoomed on the last 100 generations
%		showing the MMAE on the test set.}
%\end{figure*}

\begin{figure*}[!t]
	\noindent \begin{centering}
		\begin{tabular}{cccc}
			\includegraphics[width=0.23\textwidth]{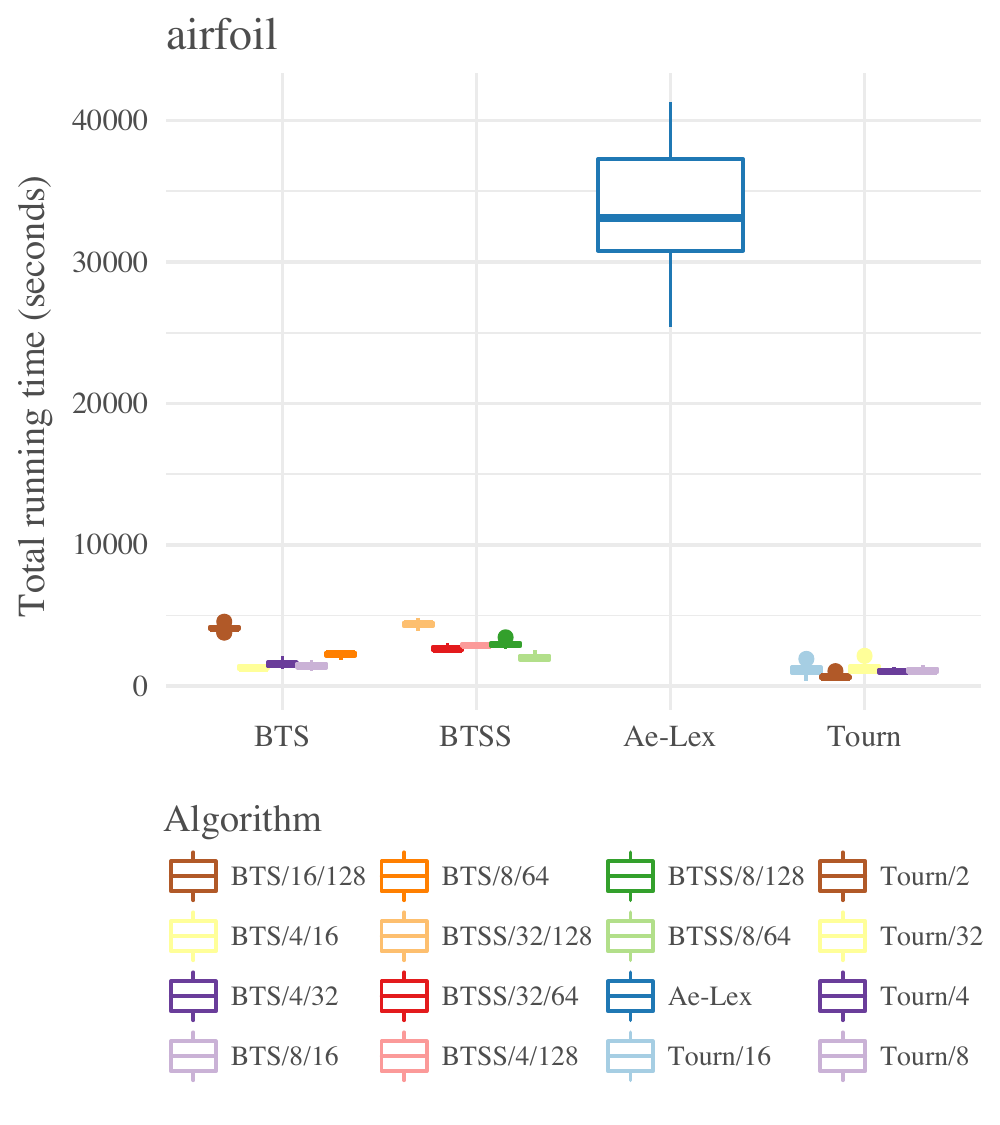} & \includegraphics[width=0.23\textwidth]{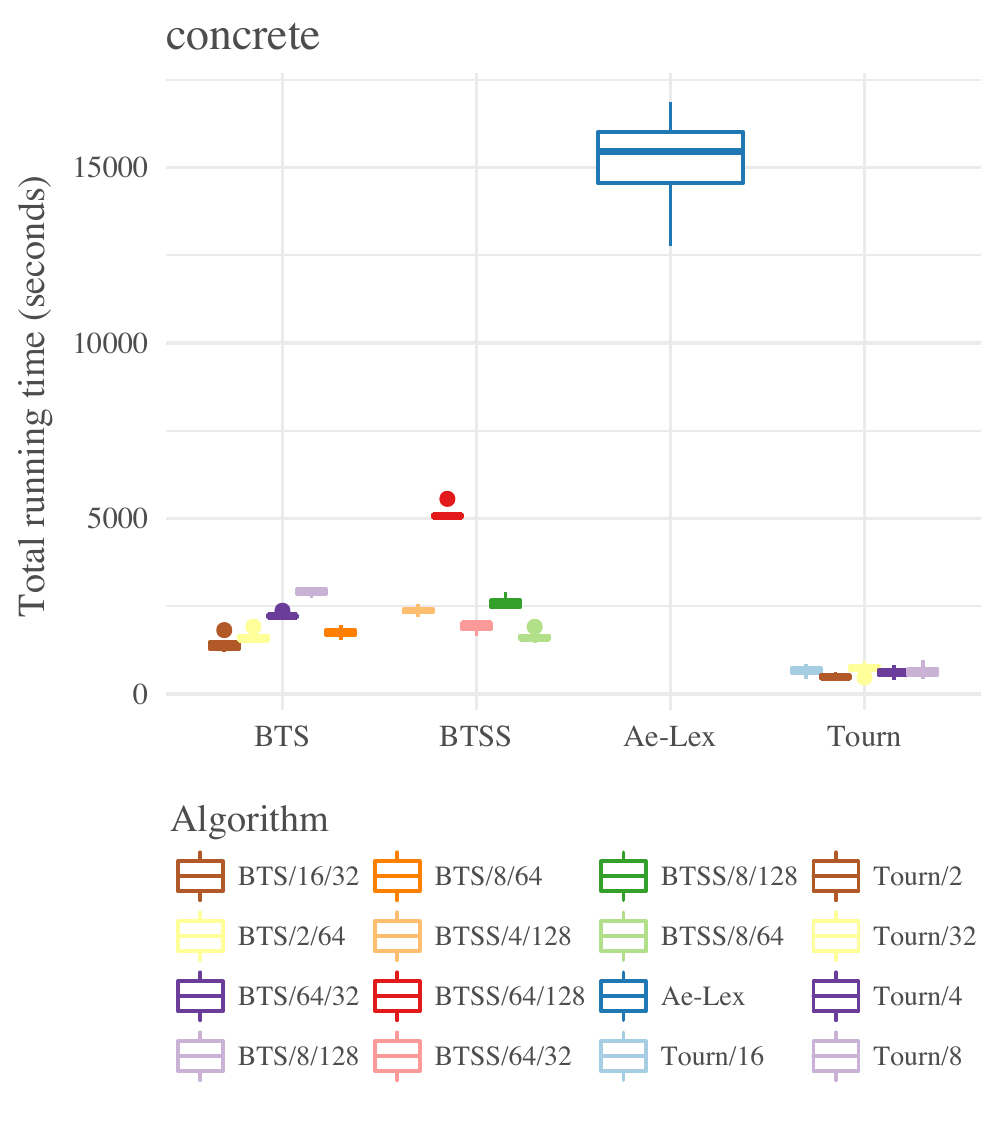} & \includegraphics[width=0.23\textwidth]{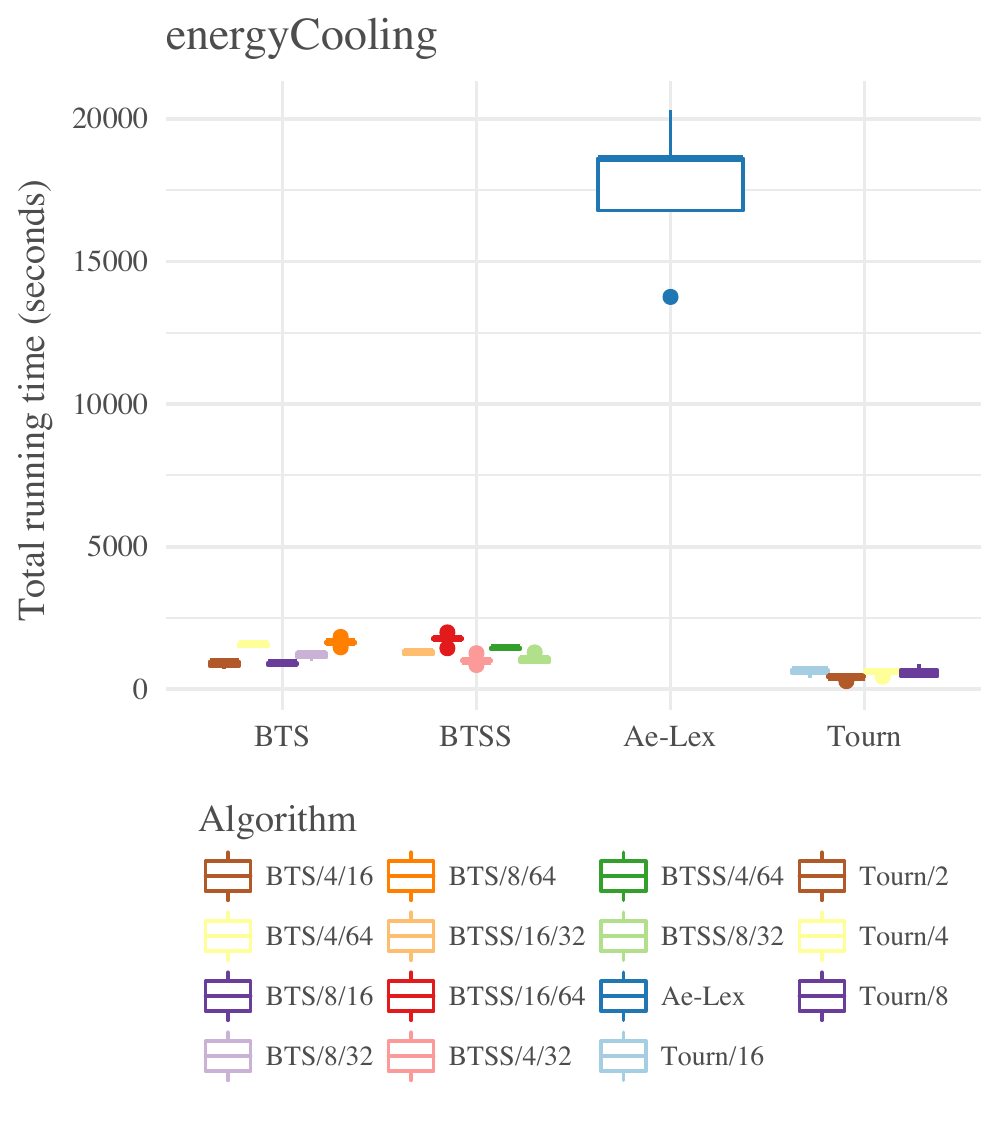} & \includegraphics[width=0.23\textwidth]{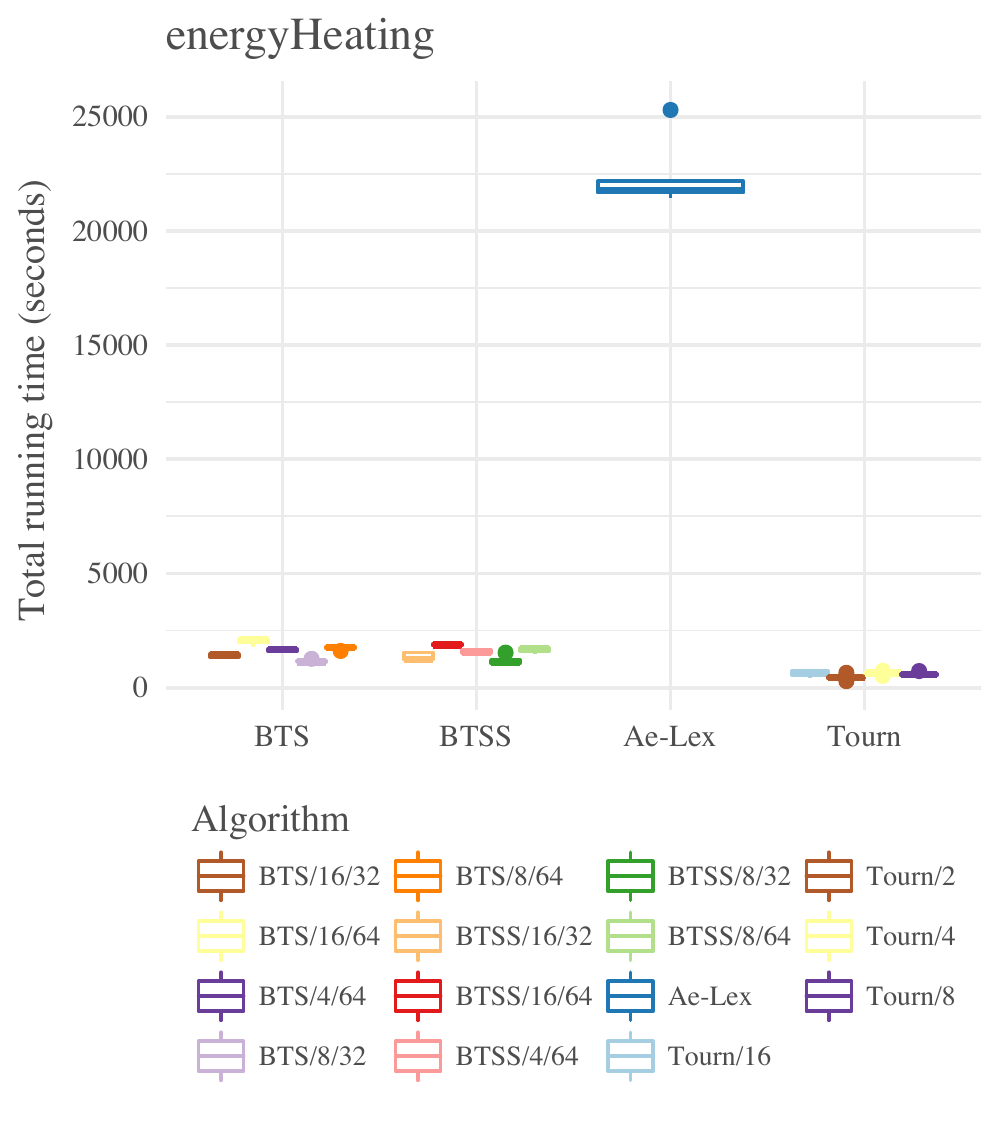}\tabularnewline
			\includegraphics[width=0.23\textwidth]{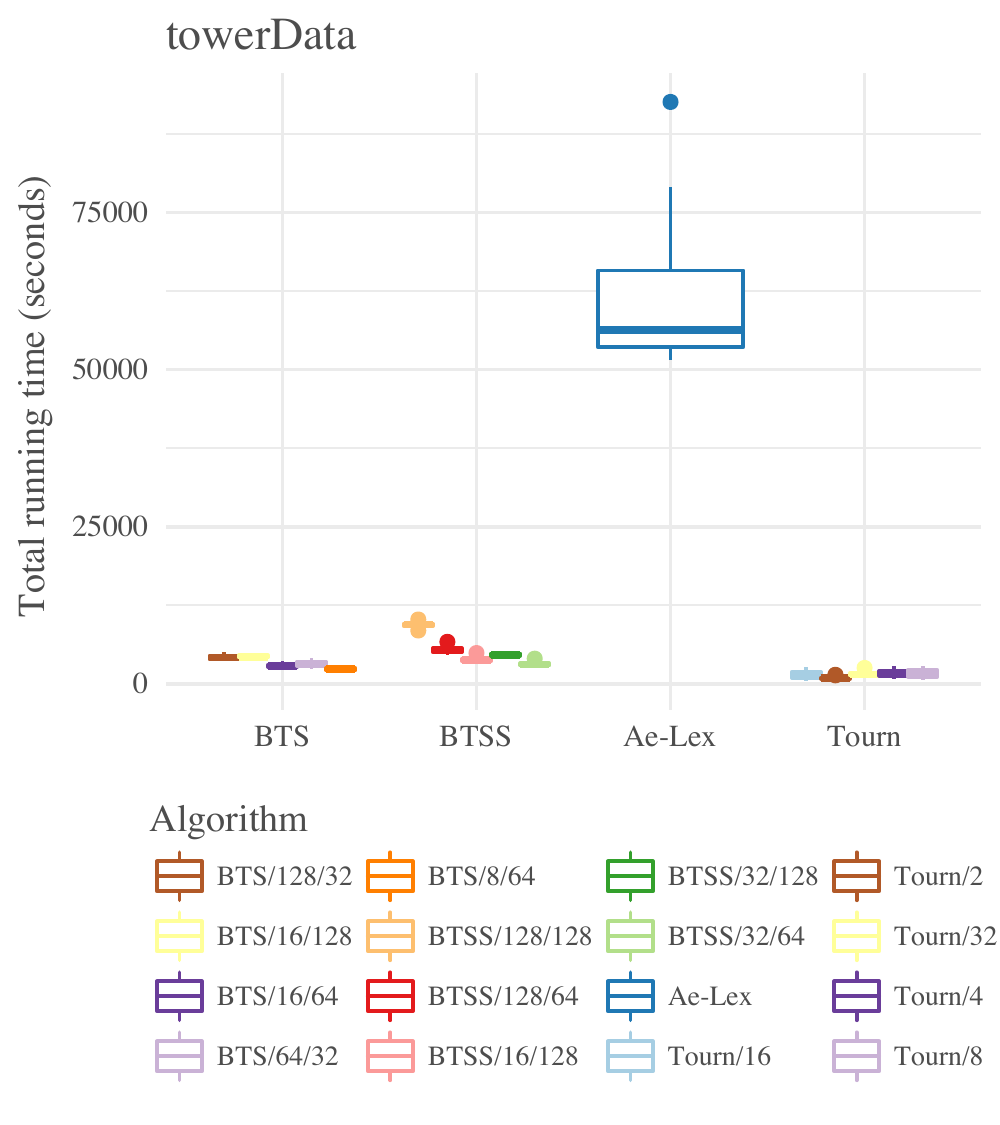} & \includegraphics[width=0.23\textwidth]{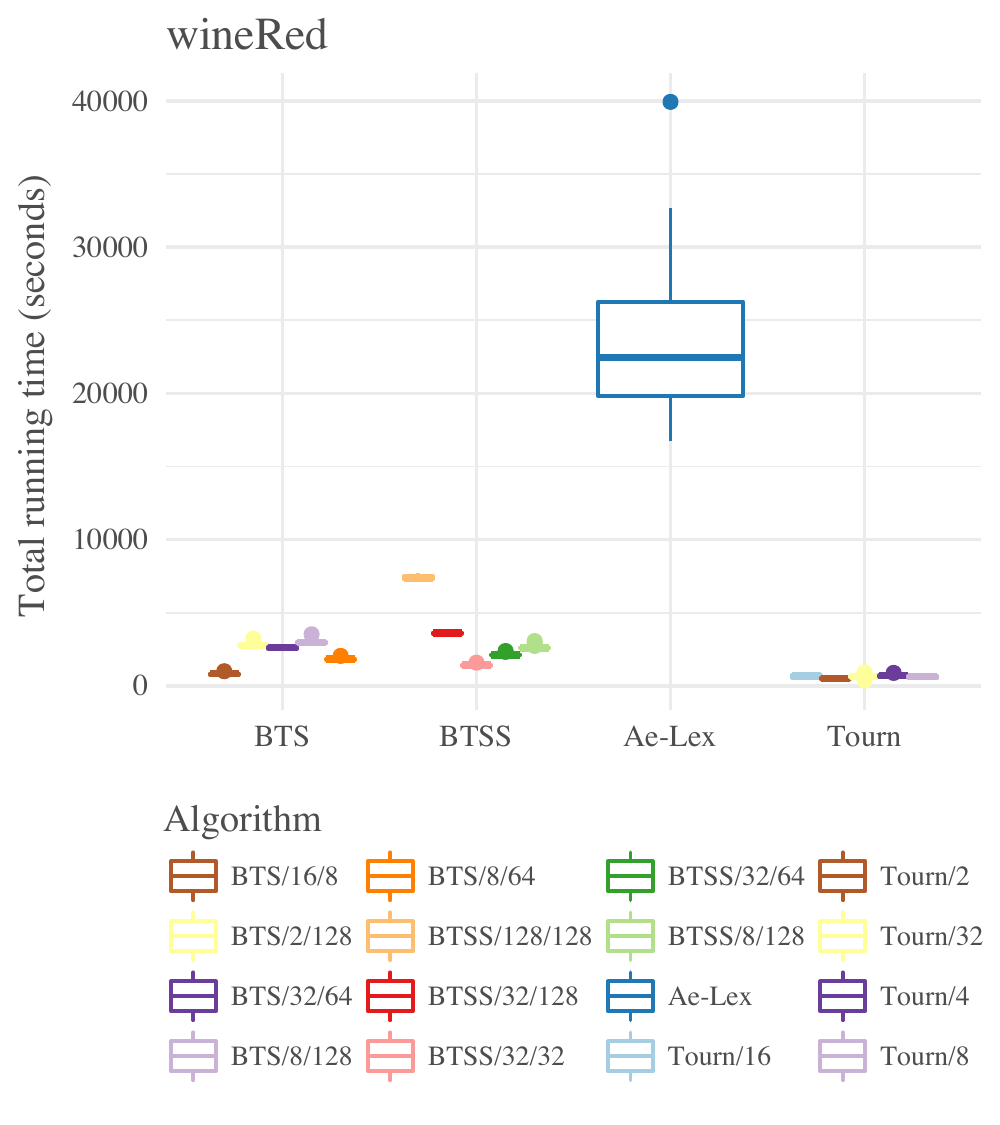} & \includegraphics[width=0.23\textwidth]{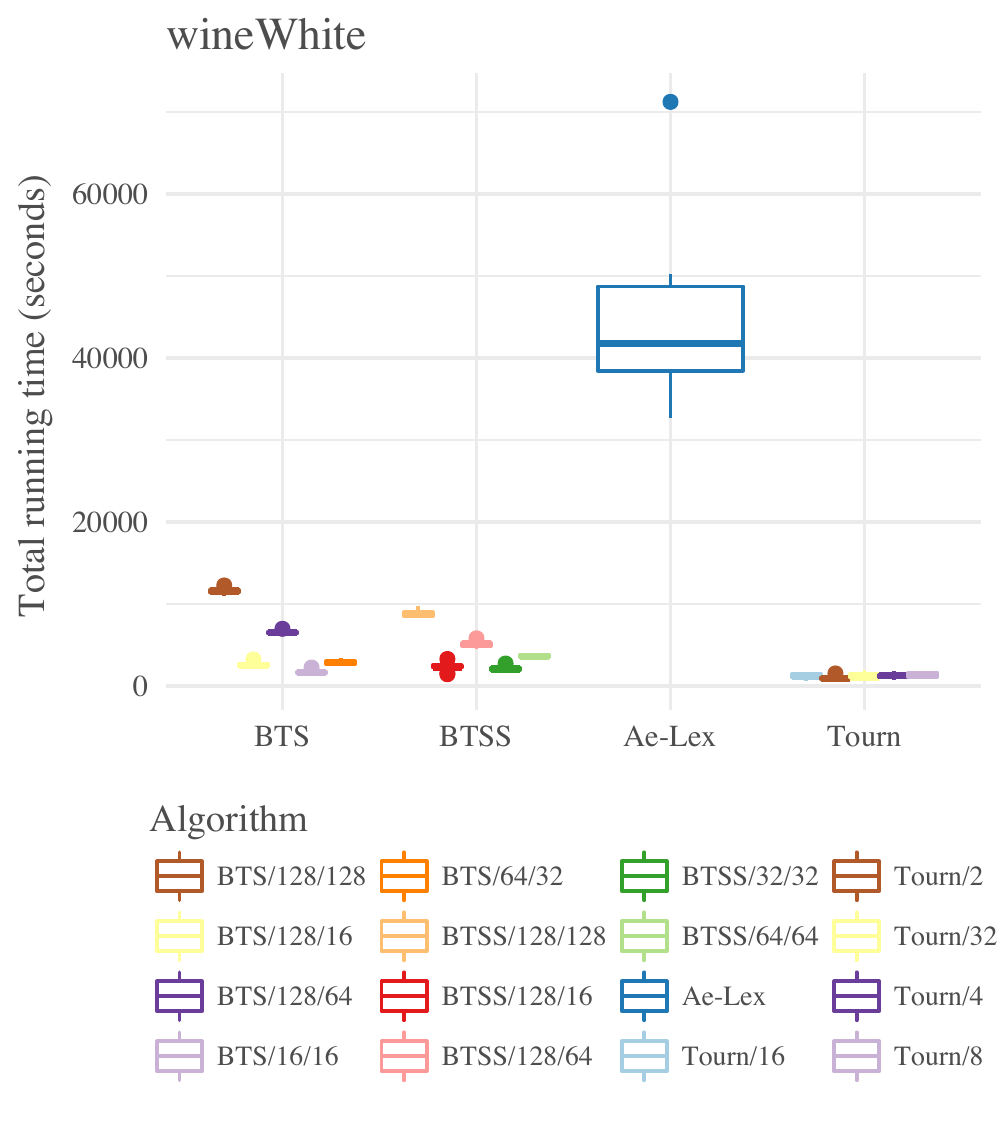} & \includegraphics[width=0.23\textwidth]{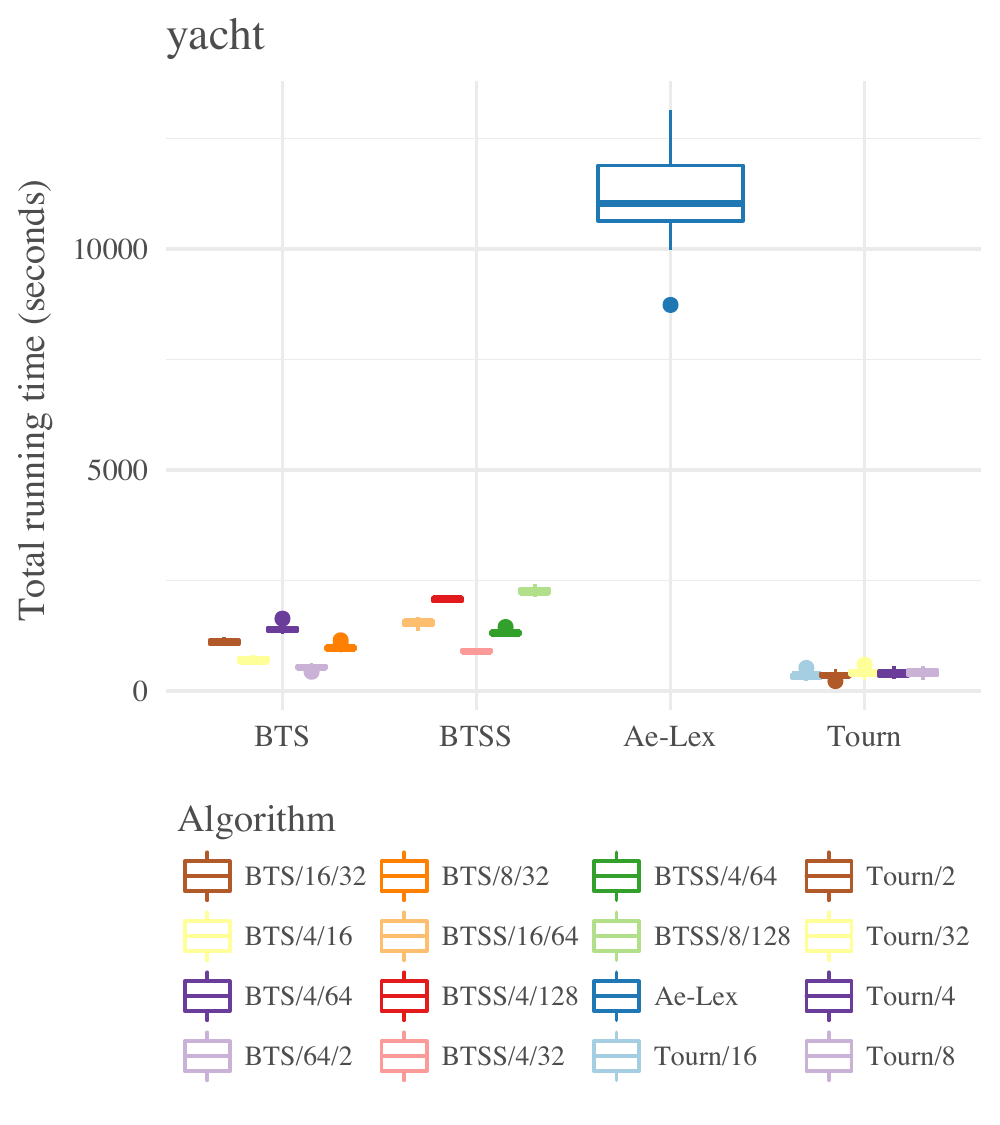}\tabularnewline
		\end{tabular}
		\par\end{centering}
	\caption{\label{fig:Boxplots-rt}Boxplots of the total running time in seconds.}
\end{figure*}

% Previsualizar código-fonte para parágrafo 53

\begin{table*}

\caption{Quality and speedup. Wilcoxon-Rank sum test ($\alpha=5\%$) with Ae-Lex
as baseline where symbol '-' means the method had lower MMAE than
Ae-Lex, and '+' means the opposite. We compare only to the Ae-Lex (the baseline).}

\noindent \centering{}\resizebox{\textwidth}{!}{%
\begin{tabular}{cccccccccccc}
\toprule 
\multicolumn{3}{c}{airfoil} & \multicolumn{3}{c}{concrete} & \multicolumn{3}{c}{energyCooling} & \multicolumn{3}{c}{energyHeating}\tabularnewline
Algorithm & MMAE & Speedup & Algorithm & MMAE & Speedup & Algorithm & MMAE & Speedup & Algorithm & MMAE & Speedup\tabularnewline
\midrule
Ae-Lex & $3.129$ & 1.00 & Ae-Lex & $5.424$ & 1.00 & Ae-Lex & $1.279$ & 1.00 & Ae-Lex & $0.733$ & 1.00\tabularnewline
BTSS/8/128 & $3.274$ & 11.30 & BTSS/64/32 & $5.223$ & 7.76 & BTSS/4/64 & $1.189$ & 12.73 & BTSS/4/64 & $0.689$ & 13.89\tabularnewline
BTSS/32/64 & $3.279$ & 12.61 & BTSS/64/128 & $5.234$ & 3.05 & BTS/8/64 & $1.237$ & 11.23 & BTSS/16/64 & $0.697$ & 11.69\tabularnewline
BTSS/8/64 & $3.286$ & 16.22 & BTSS/8/64 & $5.337$ & 9.79 & BTSS/16/32 & $1.258$ & 14.18 & BTS/16/32 & $0.740$ & 15.13\tabularnewline
BTSS/32/128 & $3.334^{+}$ & 7.55 & BTSS/8/128 & $5.350$ & 6.08 & BTS/4/64 & $1.267$ & 11.61 & BTS/8/64 & $0.769$ & 12.44\tabularnewline
BTSS/4/128 & $3.377^{+}$ & 11.46 & BTSS/4/128 & $5.370$ & 6.48 & BTS/8/16 & $1.313$ & 20.12 & BTSS/8/64 & $0.777$ & 12.83\tabularnewline
BTS/4/32 & $3.467^{+}$ & 20.77 & BTS/8/128 & $5.444$ & 5.27 & BTS/4/16 & $1.355$ & 19.22 & BTS/4/64 & $0.785$ & 13.04\tabularnewline
BTS/8/64 & $3.467^{+}$ & 14.30 & BTS/8/64 & $5.583$ & 8.72 & BTSS/8/32 & $1.370$ & 17.16 & BTSS/16/32 & $0.820$ & 17.64\tabularnewline
BTS/16/128 & $3.526^{+}$ & 8.10 & BTS/2/64 & $5.745^{+}$ & 9.85 & BTS/8/32 & $1.411$ & 14.82 & BTS/8/32 & $0.856^{+}$ & 19.32\tabularnewline
BTS/4/16 & $3.557^{+}$ & 25.84 & BTS/64/32 & $5.841^{+}$ & 6.96 & BTSS/16/64 & $1.423$ & 10.48 & BTSS/8/32 & $0.856^{+}$ & 19.86\tabularnewline
BTS/8/16 & $3.626^{+}$ & 23.63 & BTS/16/32 & $5.883^{+}$ & 11.53 & BTSS/4/32 & $1.571^{+}$ & 18.80 & BTS/16/64 & $0.940^{+}$ & 10.44\tabularnewline
\midrule 
\multicolumn{3}{c}{towerData} & \multicolumn{3}{c}{wineRed} & \multicolumn{3}{c}{wineWhite} & \multicolumn{3}{c}{yacht}\tabularnewline
Algorithm & MMAE & Speedup & Algorithm & MMAE & Speedup & Algorithm & MMAE & Speedup & Algorithm & MMAE & Speedup\tabularnewline
\midrule
Ae-Lex & $32.423$ & 1.00 & Ae-Lex & $0.472$ & 1.00 & Ae-Lex & $0.603$ & 1.00 & Ae-Lex & $0.686$ & 1.00\tabularnewline
BTS/16/128 & $27.358^{-}$ & 13.29 & BTS/8/128 & $0.449$ & 7.58 & BTSS/128/128 & $0.524^{-}$ & 4.79 & BTS/64/2 & $0.648$ & 20.78\tabularnewline
BTSS/32/128 & $27.505^{-}$ & 12.32 & BTSS/32/128 & $0.463$ & 6.22 & BTSS/64/64 & $0.529^{-}$ & 11.57 & BTS/16/32 & $0.697$ & 10.05\tabularnewline
BTSS/32/64 & $27.843^{-}$ & 18.35 & BTS/8/64 & $0.464$ & 12.26 & BTS/64/32 & $0.550^{-}$ & 14.41 & BTSS/16/64 & $0.702$ & 7.06\tabularnewline
BTSS/128/128 & $28.036^{-}$ & 5.95 & BTS/32/64 & $0.467$ & 8.59 & BTSS/128/16 & $0.560^{-}$ & 17.79 & BTS/4/16 & $0.708$ & 16.24\tabularnewline
BTS/64/32 & $28.266^{-}$ & 18.48 & BTSS/8/128 & $0.468$ & 8.67 & BTSS/128/64 & $0.570^{-}$ & 8.16 & BTSS/4/128 & $0.709$ & 5.30\tabularnewline
BTSS/128/64 & $28.275^{-}$ & 10.23 & BTSS/128/128 & $0.480$ & 3.04 & BTS/128/16 & $0.574^{-}$ & 16.81 & BTS/4/64 & $0.729$ & 7.95\tabularnewline
BTS/8/64 & $28.826^{-}$ & 23.83 & BTSS/32/32 & $0.481$ & 15.90 & BTS/128/64 & $0.592^{-}$ & 6.36 & BTSS/4/32 & $0.759$ & 12.26\tabularnewline
BTS/16/64 & $29.224^{-}$ & 19.91 & BTSS/32/64 & $0.503$ & 10.63 & BTS/128/128 & $0.593^{-}$ & 3.60 & BTSS/4/64 & $0.759$ & 8.41\tabularnewline
BTSS/16/128 & $29.594^{-}$ & 14.90 & BTS/2/128 & $0.505$ & 8.14 & BTSS/32/32 & $0.594^{-}$ & 20.05 & BTSS/8/128 & $0.765$ & 4.93\tabularnewline
BTS/128/32 & $30.142$ & 13.49 & BTS/16/8 & $0.558$ & 27.11 & BTS/16/16 & $0.611$ & 24.86 & BTS/8/32 & $0.870$ & 11.34\tabularnewline
\bottomrule
\end{tabular}}
\end{table*}

\begin{figure}
	\begin{centering}
        \includegraphics[width=.7\columnwidth]{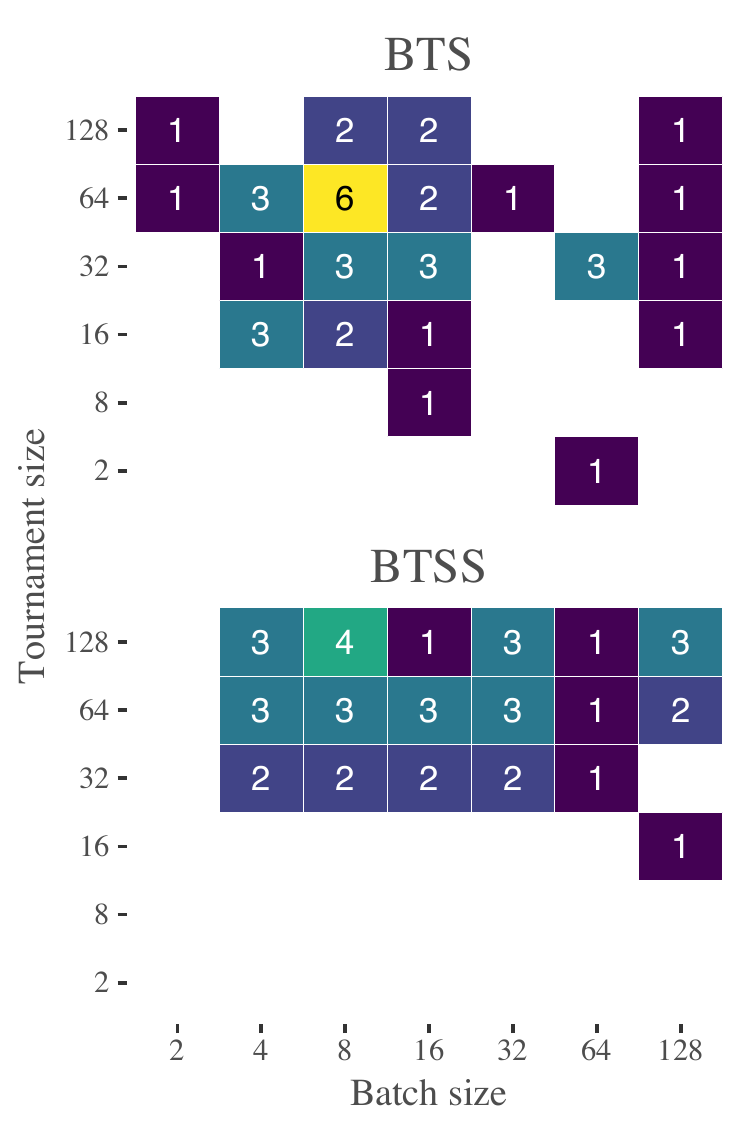}
		\par\end{centering}
	\caption{\label{fig:Top-configs} Frequency for configurations that reside in the top-five for any of the data sets. 
Notice that we do not depict a rank here but how many times each configuration was among the best. 
A blank cell means that this configuration was not in the top-five for any of the data sets. 
	}
% \end{figure*}
\end{figure}

Thus, BTS and BTSS are shown to achieve similar if not superior performance to Ae-Lex.
Interestingly, BTS and BTSS are based on a simpler computation which allows it to reach speed-ups of circa $25$ (Table~\ref{fig:Top-configs} and Figure~\ref{fig:Boxplots-rt}).
This speedup derives from the fact that both BTS and BTSS behave similarly to tournament selection and therefore have a lower computation complexity than Ae-Lex.

In Figure~\ref{fig:Top-configs}, the frequency for the top five configurations are displayed. 
Small batches that are big enough to decrease randomness but small enough to allow for specialization are usually among the best configurations.
Moreover, batches that change constantly create selection pressures that vary from generation to generation, decreasing the effects of any specialization that could take place.
This explains the slight improvement of BTS over BTSS.
In other words, in comparison with shuffled batches (BTSS), batches that are ordered by difficulty (BTS) tend to keep batches without changes throughout generations.
This allows for better specializations.
Having said that, the probability distribution of batches is constant in BTSS and therefore there is a number of batch permutations that are more probable than others.
Therefore, BTSS also allows for certain specializations to appear.

% \begin{figure*}

\begin{figure}
	\begin{centering}
		\includegraphics[width=.8\columnwidth]{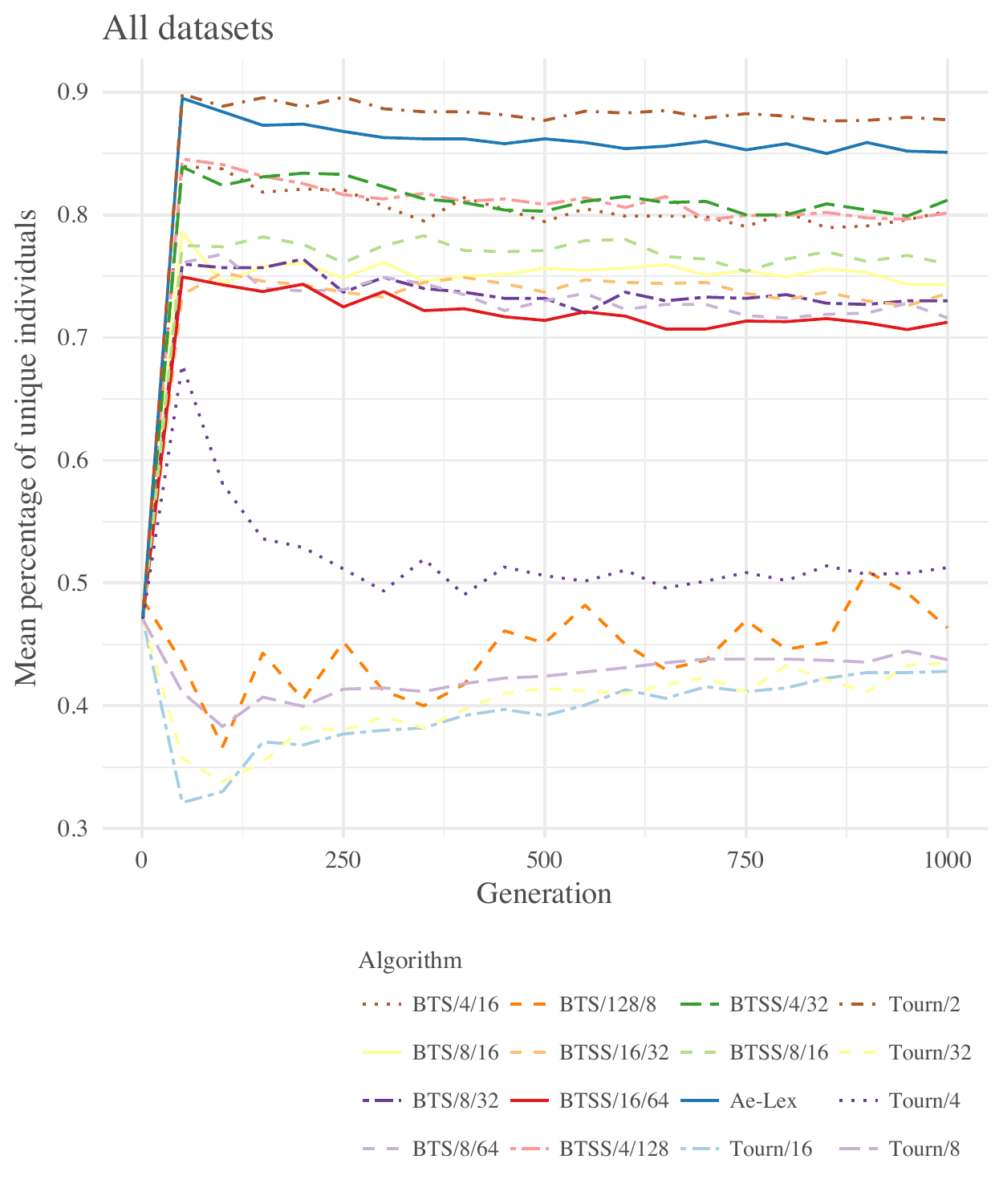}
		\par\end{centering}
	\caption{\label{fig:Curves-diversity}Diversity of the population over the generations for the top-five overall configuration.}
	\label{diver}
\end{figure}

\subsubsection{Diversity Analysis}

In this paper, we are using a straightforward diversity measure that considers only the individual's value (its fitness), not its structure. Thus, if two completely different individuals provide the same result (fitness) they are considered equal. Here we are interested in how the individuals perform, not how they look like. Structural analysis can be done in a future work.

Diversity in both lexicase and tournament selection depend on how competition takes place and both batch as well as tournament size play important roles.
To analyze deeply the diversity behavior, we did some experiments comparing some of the best performing algorithms in terms of diversity (Figure~\ref{diver}).
It is clear that pure tournaments do not reach the diversity of Ae-Lex.
This happens because in tournaments generalists are usually favoured.
Which causes the entire population to be similar 
(Tourn/2 is an exception because the size of the tournament is small enough to make the influence of the tournament almost zero.
Consequently, selection becomes mostly random. 
This naturally allows for a higher diversity with a trade-off on quality of solutions).
Regarding Ae-Lex and BTS/BTSS, around $5$ layers of diversity can be observed in Figure~\ref{diver}: $Ae-Lex >  BTSS4/32, BTSS4/128, BTS4/16 > BTSS8/16, BTS8/16 > BTS8/32, BTSS16/32, BTS8/64, BTSS16/64 > BTS128/8$.
This sequence can be explained by recalling the roles of both batch size and tournament size.
In one hand, batch size controls the overall competition scenario.
When batch size is big, batches will tend to have a similar distribution, reducing the niching effect and consequently favouring general solutions.
Small batches have the opposite effect. 
Since general solutions are similar by definition, the overall diversity of the population decreases.
This explains why bigger batch sizes result in less diversity.
In fact, all the $5$ layers shown in Figure~\ref{diver} respect this sequence.

Regarding the tournament size, it does not influence diversity strongly.
Notice that even if they allow some sub-optimal individuals to survive, these sub-optimal individuals have a small chance of surviving the following generation.
Thus, tournament can be seen mostly as an approximation to the maximum function of selecting the optimum individuals.
Having said that, very small tournament sizes may cause generalists to survive even in small batches.
This happens because the chance of a specialist individual being chosen to compete in a favourable batch is smaller.
Therefore, general solutions which perform better overall would be favoured.
This partially explains the diversity of BTS128/8 but the batch size of $128$ is also strongly affecting the diversity towards general solutions.
Moreover, Figure~\ref{diver} demonstrates that tournament size of $16$, $32$ and $128$ achieve similar diversity. 

In summary, BTS and BTSS  show that the use of batches can achieve a diversity that rivals Ae-Lex while being efficient.
Canonical tournament selection runs fail to achieve the level of diversity which further explains their poor solution quality.

\subsubsection{Batches and Ordered Test Cases}

Surprisingly, both BTS and BTSS do not use ordered test cases but batches while achieving similar accuracy and diversity to Ae-Lex.
Batches may have a totally different characteristic but both batches and test sequences allow for specialization to happen.
In test sequences, order of test cases creates a priority which fosters specialization.
In batches, groups of test cases define niches in which specialization happens.

\section{On the Survival of Specialists}

In Nature, there is no global fitness function.
Geography plays a big role in dividing species into niches.
And even in narrow places, species find ways to survive which differ from each other, creating in fact different fitness functions.
Diversity is itself a byproduct of  specialization caused by natural multi-objectivization.
In fact, diversity measures within a single population have already been shown to not help evolution, causing a deleterious conflict between diversity and selection pressure that can only be solved with sub-populations and multiple objectives \cite{vargas2015general}.

In BTS and BTSS, similarly to Ae-Lex, diversity is achieved through specialization and multi-objectivization.
Batches can be seen as different fitness functions (multi-objectivization) which create different selection pressures, allowing for diversity to appear naturally.
Specialization seems to arrive as a consequence of batches with few changes throughout generations.

\section{Conclusions}

%\vinicius{I wrote the conclusions using the current partial results. Experiments may take another week to finish but they probably won't change much.}

In this paper, we proposed the BTS, which combines the ability of lexicase selection to provide accurate solutions
with the computational complexity of simple tournament selection.
Interestingly, BTS achieves with batches  a  diversity similar to lexicase selection, showing that batches are indeed very similar in behavior to the ordering of test cases.

We argued that the similarity here derives from the fact that both BTS and lexicase selection foster  specialization of individuals.
In fact, fitness differs from batch to batch and therefore it can be said that different fitness measurements take place within a population.
Consequently, BTS allows for multi-objectiviza\-tion to appear, increasing diversity and driving specialization forward. 

%The hypothesis is that individuals exhibiting poor
%overall quality could perform well on cases of specific difficulty
%and be selected as parents, while only specialized individuals would be able
%to perform well on batches with mixed cases. After evaluating these
%methods on eight regression data sets, we found no statistical evidence
%of superiority between BTS and BTSS.

%Regarding the solution quality of Ae\_Lex and tournament selection,
%on the other hand, the hypothesis tests showed that BTS/BTSS outperformed
%Ae\_Lex on two data sets and lost on a single data set, while Tournament
%selection lost in all data sets. Thus, one can assume that BTS/BTSS
%will produce solution quality as good as that provided
%by Ae\_Lex.

%With respect to running time, BTS/BTSS was up to 25 times faster than
%Ae\_Lex, but slower than tournament selection. In terms of computational
%costs, therefore, BTS/BTSS is an order of magnitude more efficient than Ae\_Lex.
In summary, this work shows the following:
\begin{itemize}
\item \textbf{Up to $25$ Times Speed Up With Similar Performance} - BTS and BTSS provided a speed up of up to $25$ times when compared to Ae-Lex.
This is a consequence of BTS and BTSS still having the computational complexity of tournament selection while having the accuracy of Ae-Lex.
\item \textbf{The Similarity Between Batches and Ordered Test Ca\-ses} - The tests show that batches have a similar behavior to ordered test cases although the principle behind their mechanisms are completely different.
This reveals that the main reason for Lexicase's success may not be the ordered test cases but a more general principle which can be achieved in different ways.
\item \textbf{The Specialization and Multi-objectivization Hypothesis} - This work sheds light on the inner workings of lexicase selection by creating an algorithm with similar behavior through a completely different mechanism.
In fact, the similarity can be seen not only in terms of accuracy but also of diversity, demonstrating that the main principle behind Lexicase is conserved.
\end{itemize}

In summary, we propose an algorithm that can rival the state-of-the-art algorithms in performance while being one order of magnitude faster.
Moreover, investigations into the diversity and overall performance of the proposed algorithms show a surprising similarity to lexicase selection, shedding light on the main principle behind Lexicase's success.

%\begin{comment}
\begin{acks}
%\vinicius{FROM THE TEMPLATE}
%The authors would like to thank the anonymous referees for their valuable comments and helpful suggestions. 
This work was supported by JST, ACT-I Grant Number JP-50166, Japan.
%is supported by the \grantsponsor{GS501100001809}{National Natural Science Foundation of China}{http://dx.doi.org/10.13039/501100001809} under Grant No.:~\grantnum{GS501100001809}{61273304} and~\grantnum[http://www.nnsf.cn/youngscientists]{GS501100001809}{Young Scientists' Support Program}.
\end{acks}
%\end{comment}